\documentclass{article}

\usepackage[numbers, compress]{natbib} %
\usepackage[eandd,final]{neurips_2026}

\usepackage[utf8]{inputenc} %
\usepackage[T1]{fontenc}    %
\usepackage{hyperref}       %
\usepackage{url}            %
\usepackage{booktabs}       %
\usepackage{amsfonts}       %
\usepackage{nicefrac}       %
\usepackage{microtype}      %
\usepackage{wasysym}

\usepackage[table]{xcolor}  %
\usepackage{graphicx}       %
\usepackage{pmboxdraw}      %
\usepackage{siunitx}        %
\usepackage{makecell}       %
\usepackage{enumitem}       %
\usepackage{multirow}
\usepackage[normalem]{ulem} %
\usepackage{soul}
\sethlcolor{black!10}
\usepackage{subfig}

\robustify\uline
\def\Uline#1{#1\llap{\uline{\phantom{#1}}}}

\newcommand{\laionbvd}{\textsc{LAION-BVD}}
\newcommand{\bvd}[1]{\mbox{BVD-#1}}
\newcommand{\commoncrawl}{CommonCrawl}

\newcommand{\clip}{CLIP}
\newcommand{\viclip}{ViCLIP}

\newcommand{\clapmodel}{CLAP}
\newcommand{\internvid}{InternVid}

\renewcommand{\paragraph}[1]{\vspace{0.1em}\noindent\textbf{#1}.}

\newcommand{\presectionspace}{\vspace{-0.2cm}}    %
\newcommand{\postsectionspace}{\vspace{-0.2cm}}   %
\newcommand{\postsubsectionspace}{\vspace{-0.1cm}} %
\newcommand{\floatbottomspace}{\vspace{-0.3cm}}   %

\definecolor{sbblue}{HTML}{5273AD}
\definecolor{sborange}{HTML}{D78357}
\definecolor{sbgreen}{HTML}{5fA86C}
\definecolor{sbred}{HTML}{BD4C53}

\usepackage{pifont}

\usepackage[capitalise]{cleveref}      %
\crefname{figure}{Figure}{Figures}
\Crefname{figure}{Figure}{Figures}

\usepackage[breakable]{tcolorbox}
\usepackage{todonotes}
\definecolor{contribution_frame}{HTML}{5E3FBE}
\definecolor{contribution_back}{HTML}{F4F0FD}

\newtcolorbox{contributions}{
  colback=blue!5!white,
  colframe=blue!65!black,
  boxrule=0.5pt,
  arc=2pt,
  left=6pt,
  right=6pt,
  top=4pt,
  bottom=4pt,
}

\definecolor{takeaway_frame}{HTML}{223F30}
\definecolor{takeaway_back}{HTML}{EAF6ED}
\newtcolorbox{takeawaybox}{
  colback=orange!10,    %
  colframe=orange!70,   %
  boxrule=0.5pt,
  arc=2pt,
  left=6pt,
  right=6pt,
  top=4pt,
  bottom=4pt,
  title=Takeaway
}

\newcounter{takeaway}
\newif\ifneurips
\neuripstrue   %

\title{\laionbvd{}: A 13-Million-Hour Open Video Dataset for Multimodal Pre-training}

\author{%
  Andreas Hochlehnert$^{1, \ast}$ \And
  Marianna Nezhurina$^{2,3, \ast}$ \And 
  Mehdi Cherti$^{2,3}$ \AND %
  Andrej Radonjic$^{4}$ \And
  Thaddäus Wiedemer$^{5,1}$ \And
  Christoph Schuhmann$^{2}$ \And
  Romain Beaumont$^{2}$ \And
  Wieland Brendel$^{5}$ \And
  Bernhard Schölkopf$^{5}$ \AND %
  A. Sophia Koepke$^{1,6, \diamond}$ \And 
  Jenia Jitsev$^{2,3, \diamond}$ \And
  Matthias Bethge$^{1, \diamond}$ \And 
  \\ [-2ex]
  \small
  $^\ast$Shared first authors \quad $^\diamond$Shared last authors \\[1.8ex]
  $^{1}$University of Tübingen, Tübingen AI Center \quad
  $^{2}$LAION \quad
  $^{3}$JSC, FZJ \quad
  $^{4}$Wynd Labs \\[1ex]
  $^{5}$MPI for Intelligent Systems, ELLIS Institute Tübingen \quad
  $^{6}$Technical University of Munich, MCML\\[1.5ex]
\\
\url{https://projects.laion.ai/bvd/}
}
\begin{document}

\maketitle

\begin{abstract}  
    We present \laionbvd{}\footnote{LAION - Big Video Dataset}, a large-scale open video dataset for multimodal learning, which contains \emph{1.3B platform-specific video URLs} collected from \commoncrawl{}. From these, we download 80M videos with a total duration of \emph{13.5 million hours}. The dataset is designed for multimodal pre-training across the video, audio, and image modalities.
    Using content-aware scene detection, we extract clips for which we synthetically generate video and audio captions. Models trained on these data achieve competitive performance on standard video-text and audio-text benchmarks, with consistent improvements as training or model scale increases.
    Additionally, we explore video frames as an alternative source of image-text data by extracting scene-changing frames. These frames exhibit a visual distribution distinct from standard web image corpora, and models trained on this dataset achieve strong image-text retrieval performance.
    We release \laionbvd{} to the research community. It significantly expands open access to multimodal videos at an unprecedented scale.
\end{abstract}

\begin{figure}[ht!]
    \centering
    \includegraphics[width=0.49\textwidth]{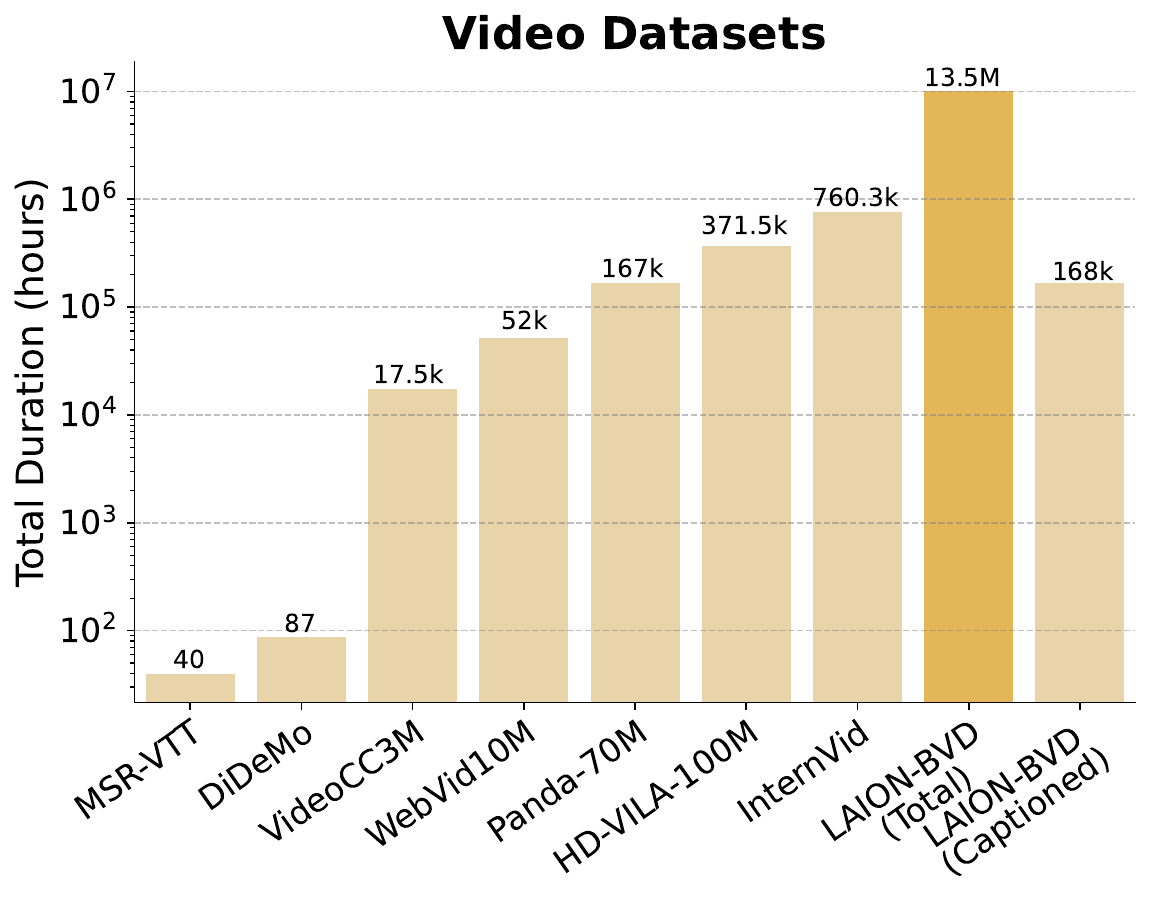}
    \hfill
    \includegraphics[width=0.49\textwidth]{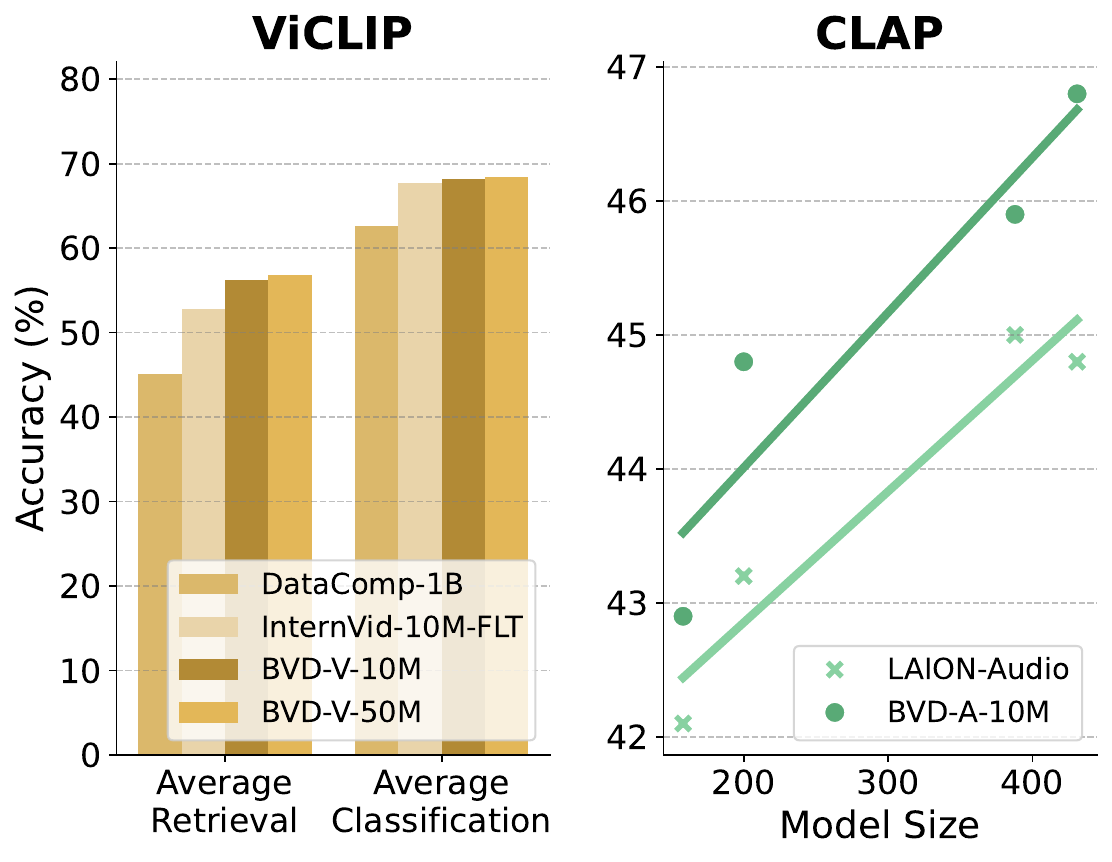}
    \caption{
        \textbf{\laionbvd{} is an open web video dataset at unprecedented scale. It facilitates strong downstream performance for multimodal pre-training.}
        \textbf{Left:} dataset scale in comparison with existing video datasets (Total: \bvd{RAW}, Captioned: \bvd{V-55M}).
        \textbf{Right:} average downstream performance of ViCLIP~\citep{wang2023internvid} (video-text, \cref{tab:exp_results_wise_ft}) and scaling trends of CLAP~\citep{laionclap2023} (audio-text, \cref{tab:exp_results_clap_pure}) on standard benchmarks.
        These results demonstrate that \laionbvd{} serves as valuable training data across both video and audio modalities.
    }
    \label{fig:teaser}
    \floatbottomspace
\end{figure}

\presectionspace
\section{Introduction}\label{sec:intro}
\postsectionspace
The current paradigm for training capable, open multimodal embedding and foundation models relies on accessible, large-scale datasets.
Many publicly available datasets exist for text~\citep{raffel2020exploring, gao2020pile, biderman2022datasheet, penedo2024the, refinedweb, weber2024redpajama} and image-text data~\citep{thomee2016yfcc100m, sharma2018conceptual, srinivasan2021wit, changpinyo2021conceptual, desai2021redcaps, schuhmann2021laion, kakaobrain2022coyo-700m, hu2022scaling, wang2023all, gadre2023datacomp, LoTLIP2024, li2024recaptionbillionswebimages, schuhmann2022laion}.
While not quite reaching the scale of proprietary datasets, they are sufficiently large to train competitive open models~\citep{schuhmann2022laion, cherti2023reproducible, Rombach_2022_CVPR, huang2023language}.
As a prominent example, LAION-5B~\citep{schuhmann2022laion} and its 2B English-language subset are among the largest public image-text datasets (with Re-LAION-5B~\citep{relaion} being their recent update) and were used to train popular models like OpenCLIP~\citep{cherti2023reproducible}, KOSMOS-1~\citep{huang2023language}, and Stable Diffusion~\citep{Rombach_2022_CVPR}.

In contrast, open video-language resources remain comparatively small.
The largest dataset to date, InternVid~\citep{wang2023internvid}, contains 7 million videos, which is relatively modest compared to modern text and image corpora. Video data can offer a particularly rich training signal, containing not only large-scale visual content but also aligned audio that can support audio-language learning.

The main bottleneck for large-scale video-text datasets is not the availability of videos, since billions are publicly accessible on the web, but the compute, storage, and engineering effort needed to download and process them. Unlike images, videos are often hosted on large commercial platforms that can restrict or gate access, making large-scale collection challenging in practice.

In this work, we address this gap with \laionbvd{}, a large-scale open web video dataset.
Starting from 1.3B platform-specific video URLs collected from \commoncrawl{} (\bvd{URLs}), we download 80M videos totaling 13.5M hours (\bvd{RAW}).
From a random sample of these, we extract scene-level clips with aligned audio as well as individual frames, and generate modality-specific captions: video captions (\bvd{V-55M}), audio captions (\bvd{A-10M}), and frame captions (\bvd{I-300M}).
To validate these subsets for multimodal pre-training, we train \viclip{}~\citep{wang2023internvid}, \clapmodel{}~\citep{laionclap2023}, and CLIP~\citep{cherti2023reproducible} models at different model and data scales.

We publish \bvd{URLs} along with the captioned subsets on \href{https://huggingface.co/collections/laion/bvd-big-video-dataset}{HuggingFace}\footnote{\url{https://huggingface.co/collections/laion/bvd-big-video-dataset}}. Furthermore, research institutions can download the raw videos (\bvd{RAW}) after accepting our \href{https://github.com/LAION-AI/BVD/blob/main/assets/bvd_terms_of_use.pdf}{terms of use}. %

Overall, our contribution can be summarized as follows:
\begin{contributions}
    \begin{itemize}[leftmargin=1em]
        \item We release \textbf{\laionbvd{}}, a large-scale open web video dataset built from 1.3B extracted video URLs (\bvd{URLs}) and 80M videos with a total duration of 13.5 million hours (\bvd{RAW}).
        \item We introduce a scalable multimodal annotation pipeline that segments videos into scenes and produces clip-level video captions, audio captions, and frame-caption pairs.
        \item Using this pipeline, we release 55M video-captioned clips (\bvd{V-55M}), 10M audio-captioned clips (\bvd{A-10M}), and 300M captioned scene-changing frames (\bvd{I-300M}) from a randomly sampled subset of \bvd{RAW}.
        \item We show that \laionbvd{} is a useful open resource for multimodal pre-training. Our \viclip{} and \clapmodel{} evaluations support video-language and audio-language learning at scale, while frame-based \clip{} experiments provide complementary evidence through strong retrieval performance.
    \end{itemize}
\end{contributions}

\presectionspace
\section{Related Work}\label{sec:related}
\postsectionspace
\paragraph{Multimodal datasets}
Training multimodal foundation models requires large-scale datasets containing data from at least two modalities.
\newlength\unitwidth

\emph{Video-text datasets} usually provide captions for clips ranging from three seconds to a few minutes~\citep{wang2023internvid, sun2024sora}.
Many early datasets are specific to domains like movies~\citep{rohrbach2017movie, soldan2022mad}, cooking~\citep{zhou2018towards, damen2018scaling}, instruction following~\citep{sanabria2018how2, miech2019howto100m}, or action recognition~\citep{soomro2012ucf101, caba2015activitynet, sigurdsson2016hollywood, kay2017kinetics, krishna2017dense, sigurdsson2018charades, goyal2017something, wang2019vatex, stroud2020learning}.
More relevant for foundation model training are general-purpose datasets that are not tied to one specific domain.
Amongst those, smaller datasets can be manually annotated~\citep{xu2016msr, hendricks2017localizingmomentsvideonatural, xu2023youkumplug10millionlargescale, liu2024valor}, but most recent and larger datasets use automatic speech recognition (ASR), subtitles, alt-text, image captions~\citep{NEURIPS2021_c6d4eb15, Xue_2022_CVPR, Bain_2021_ICCV, nagrani2022learning}, and/or utilize LLMs~\citep{wang2023internvid, wang2024swapattentionspatiotemporaldiffusions, NEURIPS2023_e6b2b48b, Chen_2024_CVPR, ju2024miradata, xiong2024lvd2mlongtakevideodataset, geng2024longvale, li2025openhumanvid, nan2025openvid, lee2021acav100m, videofactory, tan2024vidgen}. Some newer resources also broaden supervision beyond purely visual captions by incorporating audio or other multimodal signals~\citep{lee2021acav100m, liu2024valor, geng2024longvale, chen2023vast}.
\Cref{tab:video-data} summarizes video-text datasets.

\begin{table}[htp]
    \centering
        \settowidth{\unitwidth}{M}
        \newcolumntype{U}{>{\columncolor{white}[0pt][\tabcolsep]\raggedright\arraybackslash}p{\dimexpr\unitwidth}}
        \newlength\videowidth
        \settowidth{\videowidth}{\textbf{Videos}}
        \newcolumntype{V}{>{\raggedleft\arraybackslash}p{\dimexpr\videowidth-\unitwidth}}
        \newlength\durationwidth
        \settowidth{\durationwidth}{\textbf{Duration}}
        \newcolumntype{D}{>{\raggedleft\arraybackslash}p{\dimexpr\durationwidth-\unitwidth}}
        
    \caption{\textbf{\laionbvd{} compared to public open-ended video-text datasets}. \bvd{V-55M} is used in our \viclip{} and \clapmodel{} experiments.
    }
    \label{tab:video-data}
         \resizebox{\linewidth}{!}{
\newcommand{\nunit}[2]{\makebox[3.7em][r]{#1\,#2}}
\newcommand{\dur}[2]{\makebox[4.2em][r]{#1\,#2}}
\newcommand{\dashcell}{\makebox[4.2em][r]{\textendash}}

\begin{tabular}{lllcccc}
\toprule
\textbf{Dataset} & \textbf{Source} & \textbf{Captions}
& \textbf{Videos}
& \textbf{Clips}
& \begin{tabular}{c}\textbf{Duration}\\ in h\end{tabular}
& \begin{tabular}{c}\textbf{Median}\\ \textbf{Resolution}\end{tabular} \\
\midrule

MSR-VTT~\citep{xu2016msr} & YouTube & Manual & \nunit{7.2}{k} & \nunit{10}{k} & \dur{40}{} & 240p \\

DiDeMo~\citep{hendricks2017localizingmomentsvideonatural} & Flickr & Manual & \nunit{10.5}{k} & \nunit{27}{k} & \dur{87}{} & \textendash \\

LongVALE~\citep{geng2024longvale} & ACAV100M & Generated & \nunit{8.4}{k} & \nunit{105}{k} & \dur{550}{} & \textendash \\

OpenVid-1M~\citep{nan2025openvid} & Panda-70M \& Others  & Generated & \nunit{\textendash}{} & \nunit{1}{M} & \dur{2.1}{k} & \textendash \\

VidGen~\citep{tan2024vidgen} & HD-VILA-100M & Generated & \nunit{3.8}{M} & \nunit{1}{M} & \dur{3}{k} & 720p \\

LVD-2M~\citep{xiong2024lvd2mlongtakevideodataset} & YouTube + Stock footage & Generated & \nunit{2}{M} & \nunit{2}{M} & \dur{11.2}{k} & 720p \\

MiraData~\citep{ju2024miradata} & YouTube + Stock footage & Generated & \nunit{330}{k} & \nunit{330}{k} & \dur{16}{k} & 720p \\

VideoCC3M~\citep{nagrani2022learning} & YouTube & Transfer & \nunit{6.3}{M} & \nunit{10.3}{M} & \dur{17.5}{k} & \textendash \\

WebVid10M~\citep{Bain_2021_ICCV}\textsuperscript{*} & Stock footage & Alt-text & \nunit{10.7}{M} & \nunit{10.7}{M} & \dur{52}{k} & 360p \\

OpenHumanVid~\citep{li2025openhumanvid} & Films, Series, Documentaries  & Generated & \nunit{134}{k} & \nunit{52.3}{M} & \dur{70.6}{k} & 720p \\

VAST-27M~\citep{chen2023vast} & HD-VILA-100M & Generated & \nunit{\textendash}{} & \nunit{27}{M} & \dur{77}{k} & 720p \\

HowTo100M~\citep{miech2019howto100m} & YouTube & ASR & \nunit{1.2}{M} & \nunit{136}{M} & \dur{134.5}{k} & 240p \\

Panda-70M~\citep{Chen_2024_CVPR} & HD-VILA-100M & Generated & \nunit{3.8}{M} & \nunit{70.7}{M} & \dur{167}{k} & 720p \\

Koala-36M~\citep{wang2025koala} & HD-VILA-100M & Generated & \nunit{\textendash}{} & \nunit{36}{M} & \dur{172}{k} & 720p \\

HD-VG-130M~\citep{videofactory} & YouTube & Generated & \nunit{\textendash}{} & \nunit{130}{M} & \dur{181}{k} & 720p \\

ACAV100M~\citep{lee2021acav100m} & YouTube & Generated & \nunit{\textendash}{} & \nunit{100}{M} & \dur{272}{k} & 720p \\

HD-VILA-100M~\citep{Xue_2022_CVPR} & YouTube & ASR & \nunit{3.3}{M} & \nunit{103}{M} & \dur{371.5}{k} & 720p \\

YT-Temporal-180M~\citep{NEURIPS2021_c6d4eb15} & YouTube & ASR & \nunit{6}{M} & \nunit{180}{M} & \dashcell & \textendash \\

\internvid~\citep{wang2023internvid} & YouTube & Generated & \nunit{7.1}{M} & \nunit{234}{M} & \dur{760.3}{k} & 720p \\

\rowcolor{black!10}\laionbvd{} (\bvd{V-55M}) & YouTube, Vimeo, Dailymotion & Generated & \nunit{2.4}{M} & \nunit{55}{M} & \dur{168}{k} & 360p \\

\rowcolor{black!10}\laionbvd{} (\bvd{RAW}) & YouTube, Vimeo, Dailymotion & \textendash & \nunit{80}{M} & \nunit{1.8}{B}$^{\dagger}$ & \dur{13.5}{M} & 720p \\
\bottomrule
\\[-1em]
\multicolumn{7}{r}{\footnotesize $\dagger$ Estimated by extrapolating the average number of clips per video in \bvd{V-55M} to all 80M videos. We release the 80M videos unsplit, along with the captioned subsets.} \\
\multicolumn{7}{r}{\footnotesize * this dataset is now defunct} \\
\end{tabular}
   }
    \floatbottomspace
\end{table}

\emph{Audio-text datasets} are comparatively smaller, but they enable text supervision for speech, music, and ambient sounds. \cref{tab:audio-data} summarizes representative existing audio-language corpora: AudioCaps~\citep{kim2019audiocaps} and Clotho~\citep{drossos2020clotho} offer human-written captions for short general-audio clips. %
At larger scale, WavCaps~\citep{mei2024wavcaps} and AudioSetCaps~\citep{bai2025audiosetcaps} broaden coverage through weak or automatically generated captions, and audio-language pre-training efforts such as LAION-CLAP further aggregate large-scale audio-text pairs from open sources~\citep{laionclap2023}. In contrast to these dedicated audio corpora, \laionbvd{} is sourced from web videos, retaining the paired visual stream alongside the audio.

\begin{table}
    \centering
    \caption{\textbf{\laionbvd{} compared to representative publicly accessible audio-text datasets}. \bvd{A-10M} is the audio-captioned subset (drawn from \bvd{V-55M}) used in our audio-language experiments; \bvd{RAW} is the full, uncaptioned corpus.}
    \label{tab:audio-data}
    \begin{tabular}{lrrr}
    \toprule
        \textbf{Dataset} & \textbf{Videos} & \textbf{Clips} & \textbf{Duration} (in h)\\
        \midrule
        Clotho~\citep{drossos2020clotho} & \textendash & $\sim$7k & 29.1\textendash58.1 \\
        AudioCaps~\citep{kim2019audiocaps} & \textendash & $\sim$46k & 142.5 \\
        Auto-ACD~\citep{sun2023large} & \textendash & 1.5M & 4.2k \\
        LAION-Audio-630K~\citep{laionclap2023} & \textendash & $\sim$634k & $\sim$4.3k\\
        AudioSet~\citep{gemmeke2017audio} & 2.1M & 2.1M & 5.8k \\
        WavCaps~\citep{mei2024wavcaps} & \textendash & 403k & $\sim$7.6k \\
        AudioVerse~\citep{xu2026scaling} & \textendash & 1.4M & 10.1k \\
        AudioSetCaps~\citep{bai2025audiosetcaps} & \textendash & 6.1M & $\sim$17k\\
        SoundVECaps~\citep{yuan2025sound} & \textendash & 1.7M & 18.4k \\
        VAST-27M~\citep{chen2023vast} & \textendash & 27M & \textendash\\
        \rowcolor{black!10}\laionbvd{} (\bvd{A-10M}) & 2.4M & 10M & 27.5k \\
        \rowcolor{black!10}\laionbvd{} (\bvd{RAW}) & 80M & \textendash & 13.5M \\
        \bottomrule
    \end{tabular}
    \floatbottomspace
\end{table}

\emph{Image-text pairs} are easy to collect at scale, since many images on the web are captioned or come with descriptive alt-text.
As a result, open image-text datasets have grown rapidly over the past decade~\citep{thomee2016yfcc100m, sharma2018conceptual, srinivasan2021wit, changpinyo2021conceptual, desai2021redcaps, schuhmann2021laion, kakaobrain2022coyo-700m, hu2022scaling, wang2023all, gadre2023datacomp, LoTLIP2024, li2024recaptionbillionswebimages}, from just 330k English image-text pairs in MS-COCO~\citep{lin2014microsoft} to over 2B in LAION-5B~\citep{schuhmann2022laion}'s English-language subset (see~\cref{tab:image-data} for more details). 
Proprietary datasets have been scaled even further to at least 100B image-text pairs~\citep{radford2021learning, jia2021scaling, chen2022pali, pham2023combined, peng2023kosmos, dong2025scalable, wang2025scaling}. An overview of non-proprietary image-text datasets is provided in \cref{tab:image-data}.
Interleaved image-text data~\citep{alayrac2022flamingo, zhu2023multimodal, laurenccon2023obelics, huang2023language, he2023wanjuan, mckinzie2024mm1, li2024omnicorpus, futeral2024moscar, awadalla2024mint} can be scaled even further and provides more surrounding context, albeit often with weaker local alignment between modalities. In this space, OmniCorpus~\citep{li2024omnicorpus} experimented with increasing data diversity by including keyframes and video transcriptions. However, both large paired image-text corpora and interleaved web documents ultimately draw from closely related web-image distributions, and image-only scaling already shows diminishing returns on traditional benchmarks~\citep{wang2025scaling}.
In our setting, captioned video frames are therefore most relevant as an additional source of image-text supervision with a distinct visual distribution. %
\begin{table}[t]
    \centering
        \settowidth{\unitwidth}{M}
        \newcolumntype{U}{>{\columncolor{white}[0pt][\tabcolsep]\raggedright\arraybackslash}p{\dimexpr\unitwidth}}
        \newlength\imgtxtwidth
        \settowidth{\imgtxtwidth}{\textbf{Img-Txt Pairs}}
        \newcolumntype{I}{>{\raggedleft\arraybackslash}p{\dimexpr\imgtxtwidth-\unitwidth}}
    \caption{\textbf{\bvd{I-300M} compared to publicly accessible image-text datasets}. As the images are sampled from videos rather than web-crawled image pages, they complement existing datasets.}
    \label{tab:image-data}
        \resizebox{\linewidth}{!}{
        \begin{tabular}{lllI@{\,}U}
        \toprule
        \textbf{Dataset} & \textbf{Source} & \textbf{Caption Source} & \multicolumn{2}{c}{\makecell[b]{\textbf{English}\\\textbf{Img-Txt Pairs}}} \\
        \midrule
        MS-COCO~\citep{lin2014microsoft} & Flickr & Manual & 330&K \\
        Visual Genome~\citep{krishna2017visual} & MS-COCO + YFCC100M & Manual & 5.4&M \\
        YFCC100M~\citep{thomee2016yfcc100m} & Flickr & Alt-text + Title & 99&M \\
        CC3M~\citep{sharma2018conceptual} & Custom web crawl & Alt-text & 3.3&M \\
        WIT~\citep{srinivasan2021wit} & Wikipedia & Alt-text + Caption & 5.5&M \\
        CC12M~\citep{changpinyo2021conceptual} & Custom web crawl & Alt-text & 12&M \\
        RedCaps~\citep{desai2021redcaps} & Reddit & Subreddit name + Title & 12&M \\
        ALT200M~\citep{hu2022scaling} & Custom web crawl & Alt-text & 203&M\\
        COYO-300M~\citep{kakaobrain2022coyo-700m} & \commoncrawl{} & Generated & 300&M \\
        LAION-400M~\citep{schuhmann2021laion} & \commoncrawl{}~\citep{commoncrawl} & Alt-text & 413&M \\
        COYO-700M~\citep{kakaobrain2022coyo-700m} & \commoncrawl{} & Alt-text & 747&M \\
        DataComp-1B~\citep{gadre2023datacomp} & \commoncrawl{} & Alt-text & 1.4&B \\
        LAION-5B~\citep{schuhmann2022laion} & \commoncrawl{} & Alt-text & 2.3&B \\
        \rowcolor{black!10}\laionbvd{} (\bvd{I-300M}) & YouTube, Vimeo, Dailymotion & Generated & 300&M \\
        \bottomrule
        \end{tabular}
    }
    \floatbottomspace
\end{table}

\paragraph{Multimodal foundation models}
Paired image-text, video-text, and audio-text data is used to train several families of foundation models~\citep{ghosh2024exploring}. Within this landscape, \laionbvd{} is most directly relevant to video-language and audio-language representation learning, while its frame-caption pairs (\bvd{I-300M}) provide supervision for image-text training.

\emph{Embedding models} like CLIP~\citep{radford2021learning, ilharco_gabriel_2021_5143773, cherti2023reproducible}, ImageBind~\citep{girdhar2023imagebind}, and other variants~\citep{bao2022vlmo, xu2023demystifying, li2022groundedlanguageimagepretraining} learn shared cross-modal embedding spaces and are common components in multimodal systems.
VideoCLIP~\citep{xu2021videoclip} and ViCLIP~\citep{wang2023internvid} are important video counterparts, complemented by self-supervised video encoders such as VideoMAE~\citep{tong2022videomae}, while LAION-CLAP~\citep{laionclap2023} provides an audio-text setup for contrastive audio-language learning.

\emph{Multimodal LLMs} like Flamingo~\citep{alayrac2022flamingo}, BLIP~\citep{dai2023instructblipgeneralpurposevisionlanguagemodels, li2022blip, li2023blip}, LLaVA~\citep{liu2023visual, liu2024improved, liu2024llava, liu2024llavanext, zhang2024llavanextvideo}, GPT-4o~\citep{openai2024gpt4o}, MiniGPT-4~\citep{zhu2023minigpt, chen2023minigpt}, ShareGPT4V~\citep{chen2024sharegpt4v}, DeepSeek-VL~\citep{lu2024deepseek, wu2024deepseekvl2mixtureofexpertsvisionlanguagemodels} and many others~\citep{laurenccon2023obelics, chen2022pali, peng2023kosmos, bai2023qwenvlversatilevisionlanguagemodel, driess2023palm, piergiovanni2024mirasol3b, lin2023sphinx, luo2023cheap, you2023ferret, wang2024cogvlm} can take images as input and produce text as output.
Even without an explicit temporal dimension during training, some image-text trained multimodal LLMs exhibit good video understanding~\citep{kim2024image}.
Other multimodal LLMs like Llama model variants~\citep{zhang2023video, li2024llama}, InternVideo-Next~\citep{wang2025internvideonext}, Qwen3-VL~\citep{qwen3vl}, Video-ChatGPT~\citep{maaz2023video}, PandaGPT~\citep{su2023pandagpt}, and others~\citep{zhang2024llavanextvideo, liu2024valor, lyu2023macaw, yan2022videococa, zhao2023chatbridge} are specifically trained with video data.

\emph{Large multimodal models} like recent Gemini~\citep{team2024gemini} models, CoDi~\citep{tang2023any, tang2024codi}, NExT-GPT~\citep{wu2024next}, and VideoPoet~\citep{kondratyuk2023videopoet} handle images and videos as input and output.

\paragraph{Image, video, and audio captioning}
Image and video captioning have a long history in deep learning, both as benchmark tasks for perceptual understanding and as tools for summarization and abstraction~\citep{vinyals2016show, you2016image, gu2017non, sharma2018conceptual, guo2020non, sidorov2020textcaps}.
Audio captioning benchmarks such as AudioCaps~\citep{kim2019audiocaps} and Clotho~\citep{drossos2020clotho} extend this to non-visual modalities, supporting the audio-text retrieval task and benchmark introduced in~\citep{oncescu2021audio,koepke2022audio}, while open audio-language models such as Audio Flamingo 3 build on this paired data for audio understanding~\citep{goel2025audio}.

Automatic captioning pipelines for multimodal data curation at scale commonly follow a shared approach~\citep{LoTLIP2024, li2024recaptionbillionswebimages, wang2023internvid, xue2024progress, Chen_2024_CVPR, geng2024longvale}:
Images (including the center frame for videos) are captioned using a strong pretrained multimodal LLM.
Videos are first split into short clips and might be annotated by an existing video-language model like LLaVA-NeXT-Video~\citep{zhang2024llavanextvideo} or frame-by-frame by a more lightweight model like Tag2Text~\citep{huang2023tag2text}.
Audio captions from models like Qwen-Audio / Qwen-Omni~\citep{chu2023qwen,Qwen3-Omni} or transcriptions from models like Whisper-Large~\citep{radford2023robust} can also be incorporated.
Final video captions are generated either by synthesizing the partial captions with pretrained language models such as T5~\citep{raffel2020exploring}, Vicuna~\citep{vicuna2023}, Gemini~\citep{team2024gemini}, and Claude~\citep{anthropic2024claude3}, or by directly captioning videos with multimodal models~\citep{yang2025vimix}. 

\presectionspace
\section{Dataset}\label{sec:method}
\postsectionspace

\begin{figure}[ht]
    \centering
    \includegraphics[width=\textwidth]{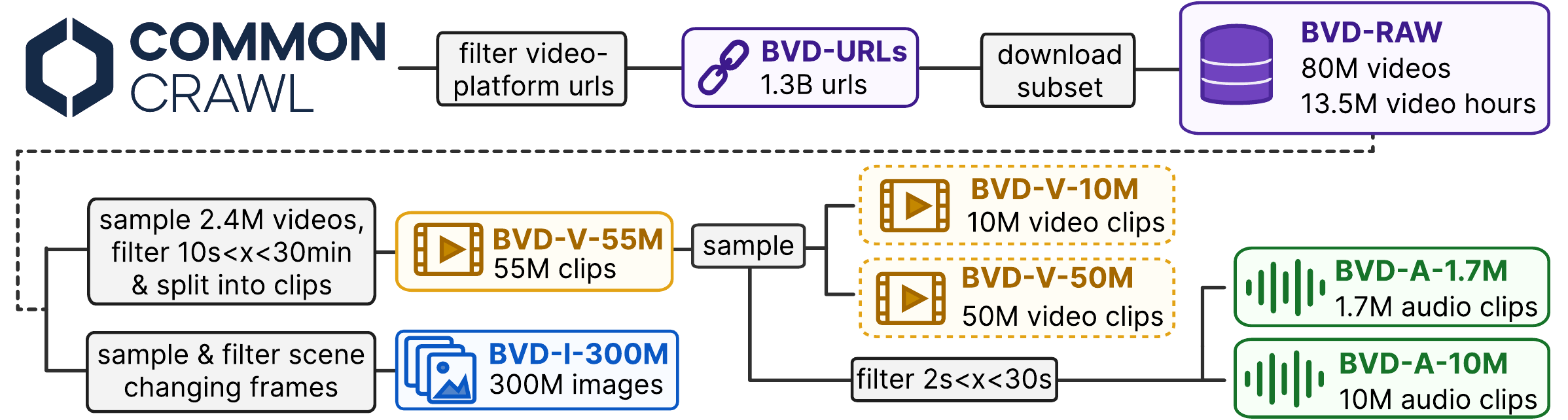}
    \caption{\textbf{\laionbvd{} data curation pipeline and subsets}. We source data from \commoncrawl{}~\citep{commoncrawl} and filter for platform-specific video URLs, resulting in 1.3B video candidates (\bvd{URLs}). Out of these, we downloaded 80M videos (13.5M hours; \bvd{RAW}) using a distributed infrastructure. From \bvd{RAW}, we derive the captioned video (\bvd{V-*}), audio (\bvd{A-*}), and frame (\bvd{I-300M}) subsets used in our experiments.}
    \label{fig:data_curation}
    \floatbottomspace
\end{figure}

In this section, we outline the data collection process for \laionbvd{} and provide key dataset statistics. Specifically, \cref{sec:curation} details the data curation procedure, \cref{sec:captioning} describes the captioning pipeline, and \cref{sec:stats} presents the dataset statistics. \Cref{fig:data_curation} gives an overview of all subsets. %
\subsection{Data curation}\label{sec:curation}

\begin{figure}[t]
    \centering
    \includegraphics[width=\textwidth]{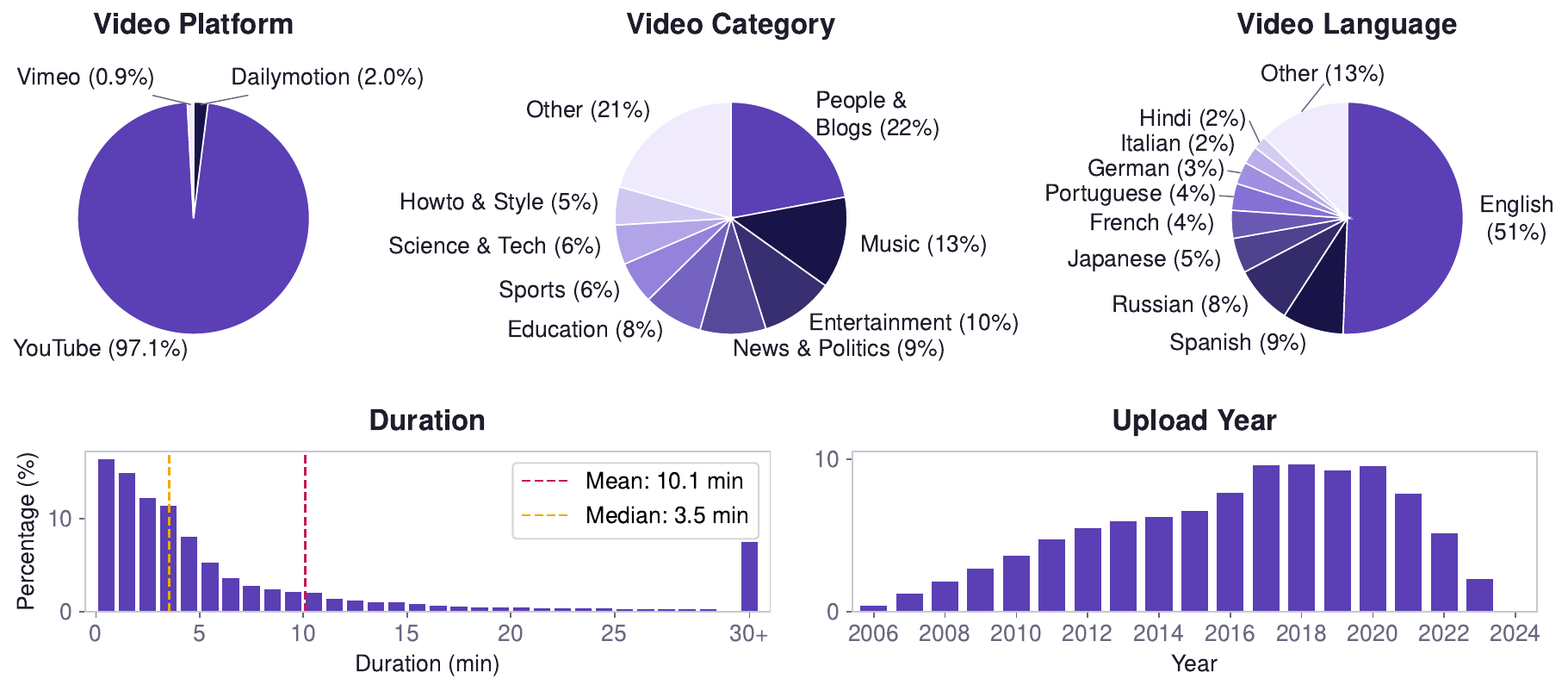}
    \caption{
        \textbf{Dataset statistics for the 80M downloaded videos (\bvd{RAW})}.
        \textbf{Top row}: Videos are sourced from YouTube, Dailymotion, and Vimeo and cover a diverse set of topics. Among the \SI{58.2}{\%} of videos with language annotations, English accounts for approximately half, with the remainder spanning a variety of other languages.
        \textbf{Bottom row}: The mean video duration is 10.1 minutes and videos cover almost two decades.
    }
    \label{fig:stats}
    \floatbottomspace
\end{figure}

\begin{figure}[t]
    \centering
    
    \includegraphics[width=\linewidth]{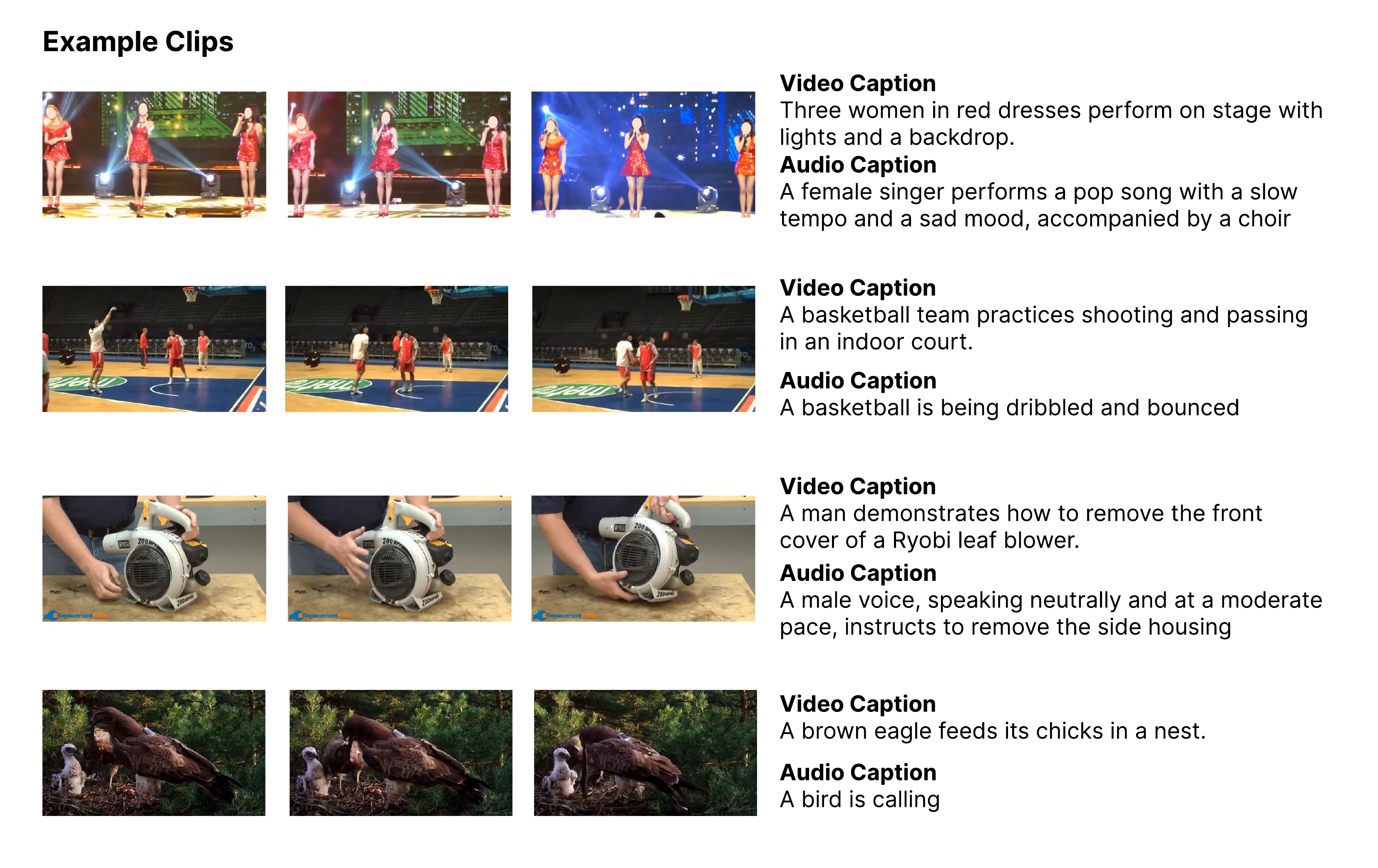}
    \vspace{0.3em}
    \includegraphics[width=\linewidth]{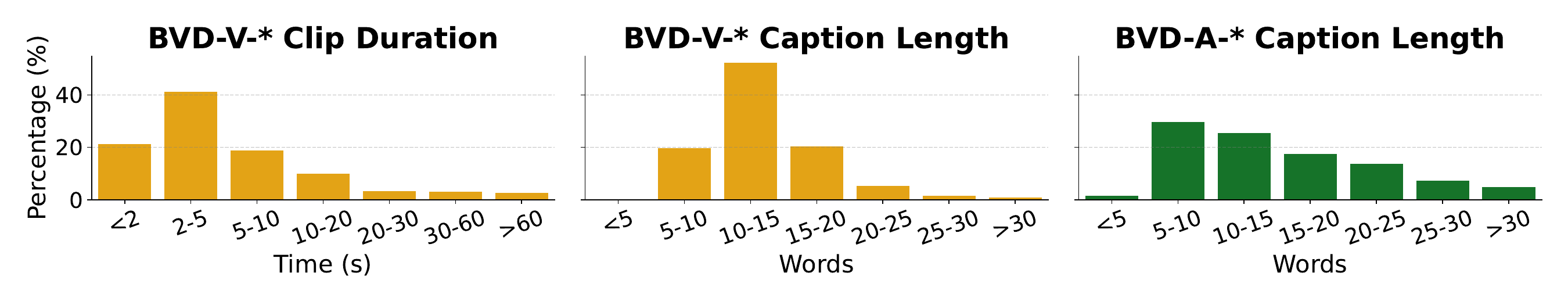}
    \caption{
        \textbf{Overview of the captioned clips in \bvd{V-55M} and \bvd{A-10M}.}
        \textbf{Top:} Sampled clips with corresponding video and audio captions.
        \textbf{Bottom:} Distribution of clip and caption lengths for the video (\bvd{V-*}) and audio (\bvd{A-*}) subsets. As the smaller subsets are uniform random samples, they follow the same distributions. %
    }
    \label{fig:examples}
    \floatbottomspace
\end{figure}

\paragraph{Sources}
Large-scale multimodal datasets like LAION-5B~\citep{schuhmann2022laion}/Re-LAION-5B~\citep{relaion} and DataComp-1B~\citep{gadre2023datacomp} rely on \commoncrawl{}~\citep{commoncrawl} as a source of raw data. Following this approach, we use \commoncrawl{} WAT (Web Archive Transformation) files, which provide essential metadata about archived web pages, including HTTP headers and hyperlinks. We extract platform-specific video links using \texttt{yt-dlp}~\citep{yt-dlp} extractors. To efficiently process this data at scale, we employ the \texttt{cc2dataset}~\citep{cc2dataset} tool in conjunction with an Apache Spark~\citep{10.1145/2934664} cluster. This setup enables the rapid extraction of video URLs and their associated metadata. We use all \commoncrawl{} dumps available as of March 2024, resulting in a corpus of 4.7B candidate URLs. To ensure the quality and accessibility of the videos, we filter for links from major supported platforms---YouTube~\citep{youtube}, Vimeo~\citep{vimeo}, and Dailymotion~\citep{dailymotion}---yielding a final set of 1.3B video URLs (\bvd{URLs}). %
We did not apply additional safety filters at the data collection stage, as these platforms are moderated for harmful and unsafe content.

\paragraph{Video download}
We download videos using a distributed setup of 2,000 virtual servers coordinated via a cluster built on \texttt{Celery}~\citep{celery} and powered by \texttt{yt-dlp}~\citep{yt-dlp}. To ensure robust access to video content, we employ a residential proxy network throughout the download process. 
Overall, we attempted to download 130M videos and achieved a link success rate of approximately \SI{60}{\%}, resulting in 80M successfully retrieved videos with a total duration of 13.5M hours (\bvd{RAW}).

\paragraph{Clip extraction for video and audio training}
We randomly sample 2.4M videos from \bvd{RAW} and extract 55M clips (\bvd{V-55M}) using a shared pre-processing pipeline for both video-language and audio-language experiments. The pre-processing pipeline is as follows. First, we discard videos shorter than 10 seconds or longer than 30 minutes. We then apply content-based scene detection using \texttt{PySceneDetect}~\citep{pyscenedetect} with a threshold of 30 and split each video into scene-level segments. To further improve clip quality, we remove segments that are effectively static by estimating frame-to-frame motion on a low-resolution sample and discarding clips with no meaningful visual change. These clips are captioned for the \viclip{} experiments, while for \clapmodel{} we caption the audio of 10M random clips of 2--30 seconds (\bvd{A-10M}; see \cref{fig:examples} for examples).

\paragraph{Frame filtering for image training}
From \bvd{RAW}, we randomly sample videos and extract keyframes using \texttt{ffmpeg}. We then filter out black frames and retain scene-change frames using \texttt{ffmpeg}'s scene detection with a threshold of $0.1$. We continue this process until reaching approximately 300M filtered frames, resulting in the \bvd{I-300M} subset.

\subsection{Captioning pipeline}\label{sec:captioning}
\postsubsectionspace

Following the download and pre-processing described above we annotate the resulting clips and frames with modality specific captioners.

\paragraph{Video captioning} We caption the \bvd{V-55M} clips using Qwen3-VL-2B-Instruct~\citep{qwen3vl} served through vLLM~\citep{kwon2023efficient}. For each clip, we sample up to 32 frames uniformly across the clip and provide them to the model together with the prompt: ``Describe the video in 20 words or less.'' The resulting output is used as a video caption for the clip (see \cref{fig:examples} for caption statistics).

\paragraph{Audio captioning} We caption audio clips with Audio Flamingo 3~\citep{goel2025audio}. For each audio segment, we prompt the model with: ``Describe the audio sounds in 10 words or less.'' This is motivated by the caption length in the AudioCaps~\citep{kim2019audiocaps} and Clotho~\citep{drossos2020clotho} audio-text datasets, which average around 8 and 11 words respectively (see \cref{fig:examples} for caption statistics).

\paragraph{Frame recaptioning for image-text analysis} To preserve comparability with existing web-image analyses, we additionally recaption sampled keyframes with DeepSeek-VL2-tiny~\citep{wu2024deepseekvl2mixtureofexpertsvisionlanguagemodels}. These captions are used for the image-text analysis. A detailed account on the caption pipeline and model selection can be found in \cref{sec:clip_captioning}.

\subsection{Dataset statistics}\label{sec:stats}
\postsubsectionspace

With 80M videos and 13.5 million video hours, \bvd{RAW} provides a large corpus of audio-visual content.
\cref{fig:teaser} and \cref{tab:video-data,tab:audio-data} contextualize our dataset within existing video- and audio-language datasets. Statistics of \bvd{RAW} are provided in \cref{fig:stats}.
Consistent with prior video-language datasets, the vast majority of videos (\SI{97.1}{\%}) are sourced from YouTube, with much smaller fractions from Dailymotion (\SI{2.0}{\%}) and Vimeo (\SI{0.9}{\%}).

The videos cover a broad range of topics, with People \& Blogs (\SI{22}{\%}), Music (\SI{13}{\%}), and Entertainment (\SI{10}{\%}) representing the largest categories, alongside a broad distribution of other content.
The dataset is also strongly multilingual.
Of the \SI{58.2}{\%} of videos with language tags, about half are in English, followed by Spanish (\SI{9}{\%}), Russian (\SI{8}{\%}), and Japanese (\SI{5}{\%}), among many others.

Like other video datasets, most videos in \bvd{RAW} are below 5 minutes long (median 3.5 minutes), with a mean duration of 10.1 minutes.
However, the distribution is rather long-tailed, and almost \SI{7.5}{\%} of videos are 30 minutes or longer, supporting long-horizon video tasks.
The collection includes relatively recent uploads, with the latest entries from early 2024.

For our modality-specific analyses, we create two subsets from \laionbvd{} as described in \cref{sec:curation,sec:captioning}:

\paragraph{Captioned video/audio subsets} Our video--text (\viclip{}) and audio--text (\clapmodel{}) data validation experiments (\cref{sec:experiments}) use captioned subsets of \bvd{V-55M} (\cref{fig:data_curation}). \bvd{V-55M} contains 55M scene-level video clips from 2.4M source videos, totaling 168k hours (mean clip duration 11\,s). \bvd{A-10M} contains 10M audio clips randomly sampled from \bvd{V-55M}, totaling 27.5k hours (mean clip duration 9.9\,s). \Cref{fig:examples} summarizes clip-duration and caption statistics and shows representative examples.

\paragraph{Captioned frame subset} \bvd{I-300M} consists of 300M captioned scene-changing keyframes sampled from \bvd{RAW}. As an additional image-centric analysis, we quantify distributional differences between web-image data and our video-derived frames by computing the Fréchet distance (FD) over CLIP-ViT-B/32 embeddings using 100k randomly sampled images from each source. We observe a substantial shift between the distributions of Re-LAION and \bvd{I-300M} frames (FD = 33.92), while two independent 100k samples from Re-LAION yield a near-zero FD (0.16). This confirms that video frames from \laionbvd{} provide a complementary visual distribution to existing web-scale image datasets, supporting its value as an additional pre-training source.

\presectionspace
\section{Experiments validating \laionbvd{}}\label{sec:experiments}
\postsectionspace
We evaluate \laionbvd{} on three modalities: video, audio, and static frames. For video and audio, we study whether the captioned clips in \bvd{V-*} and \bvd{A-*} (\cref{fig:data_curation}) support video-text and audio-text learning with \viclip{} and \clapmodel{}, respectively. For static frame analysis, we assess video frames as a source of image-text data by training \clip{} on \bvd{I-300M}. Hyperparameters and experimental details, are provided in \cref{sec:experimental_details}.

\subsection{Video-text data validation with \viclip{}}\label{sec:viclip_results}
\postsubsectionspace

For training \viclip{}, we use \bvd{V-55M} (\cref{sec:curation}). Following~\citep{wang2023internvid}, we further construct two smaller video subsets of 10M and 50M samples, sampled uniformly from \bvd{V-55M}, which we denote as \bvd{V-10M} and \bvd{V-50M}. We evaluate video performance on three zero-shot action recognition benchmarks (Kinetics-400~\citep{kay2017kinetics}, UCF-101~\citep{soomro2012ucf101}, and HMDB51~\citep{kuehne2011hmdb}) and two video-text retrieval datasets (MSR-VTT~\citep{xu2016msr} and MSVD~\citep{chen2011collecting}), reporting a total of three classification metrics (top-1 accuracy) and four retrieval metrics (video-to-text retrieval recall@1 and text-to-video retrieval recall@1). 
To provide a concise summary of these results, we report an overall average (denoted \textbf{Avg} in \cref{tab:exp_results_viclip_scaling}) defined as the arithmetic mean of two components: the average classification accuracy across the three classification metrics and the average of the four retrieval metrics across both retrieval datasets.
\definecolor{bvdred}{HTML}{CA498C}
\definecolor{bvdviolet}{HTML}{5E3FBE}

\begin{table}[t]
\centering
\small
\caption{\textbf{ViT-L/14 scaling with \bvd{V-*} data on \textcolor{bvdred}{classification} and \textcolor{bvdviolet}{retrieval} tasks}.
\viclip{} trained on \bvd{V-10M} and \bvd{V-50M} outperforms both InternVid-trained \viclip{} and the image-only CLIP baseline trained on DataComp-1B where video embeddings are obtained by averaging frame embeddings. Performance consistently improves with increased training data and dataset scale, with \bvd{V-50M} achieving the best overall average.}
\label{tab:exp_results_viclip_scaling}
\resizebox{\textwidth}{!}{%
\begin{tabular}{l c c c c c c c c r}
\hline
Dataset & Samples & \textcolor{bvdred}{\textbf{K400}} & \textcolor{bvdred}{\textbf{UCF-101}} & \textcolor{bvdred}{\textbf{HMDB51}} & \multicolumn{2}{c}{\textcolor{bvdviolet}{\textbf{MSR-VTT}}} & \multicolumn{2}{c}{\textcolor{bvdviolet}{\textbf{MSVD}}} & \textbf{Avg} \\

 & seen & top-1 & top-1 & top-1 & T2V & V2T & T2V & V2T & \\
\hline
DataComp-1B (image only) & {-} & 61.2 & 78.1 & 48.1 & 37.3 & 26.9 & 48.3 & 63.3 & 53.2 \\
InternVid-10M* & 10M & 56.7 & {-} & {-} & 36.4 & 37.1 & 45.2 & 69.8 & {-} \\
InternVid-50M* & 50M & 57.2 & {-} & {-} & 39.7 & 40.7 & 46.5 & 72.2 & {-} \\
InternVid-10M-FLT* & 10M & \textbf{64.8} & {-} & {-} & 42.4 & 41.3 & 49.1 & 75.1 & {-} \\
InternVid-10M-FLT & 10M & 61.8 & 78.6 & 55.1 & 38.7 & 39.8 & 50.9 & 74.1 & 58.0 \\
\rowcolor{gray!10}\bvd{V-10M} & 10M & 62.7 & 79.5 & 60.4 & 42.7 & 42.9 & 53.2 & 81.0 & 61.3 \\
\rowcolor{gray!10}\bvd{V-10M} & 50M & 63.1 & 79.4 & 60.3 & 41.3 & 43.1 & 53.2 & 83.0 & 61.4 \\
\rowcolor{gray!10}\bvd{V-50M}  & 10M & 63.3 & 78.9 & 60.4 & \textbf{43.5} & 42.6 & 53.6 & 82.1 & 61.5 \\
\rowcolor{gray!10}\bvd{V-50M} & 50M & 63.3 & \textbf{79.7} & \textbf{61.4} & 42.8 & \textbf{43.2} & \textbf{53.7} & \textbf{83.9} & \textbf{62.0} \\
\hline
\multicolumn{10}{r}{\footnotesize * results reported in \citep{wang2023internvid}; all other results are evaluated using our pipeline.} \\
\end{tabular}
}
    \floatbottomspace
\end{table}

\begin{table}[t]
\centering
\small
\caption{%
\textbf{Scaling behavior of \viclip{} on \bvd{V-55M} across \textcolor{bvdred}{classification} and \textcolor{bvdviolet}{retrieval} tasks}. Results demonstrate consistent performance gains with increased model size and samples seen.
}
\label{tab:exp_results_viclip_ref}
\resizebox{\textwidth}{!}{%
\begin{tabular}{l l c c c c c c c c r}
\hline
Model & Dataset & Samples & \textcolor{bvdred}{\textbf{K400}} & \textcolor{bvdred}{\textbf{UCF-101}} & \textcolor{bvdred}{\textbf{HMDB51}} & \multicolumn{2}{c}{\textcolor{bvdviolet}{\textbf{MSR-VTT}}} & \multicolumn{2}{c}{\textcolor{bvdviolet}{\textbf{MSVD}}} & \textbf{Avg} \\
& & seen & top-1 & top-1 & top-1 & T2V & V2T & T2V & V2T & \\
\hline

ViT-B/32 & \bvd{V-55M} & 10M & 49.9 & 70.2 & 44.6 & 34.2 & 38.1 & 44.2 & 74.8 & 51.4 \\
ViT-B/32 & \bvd{V-55M} & 50M & 51.6 & 70.4 & 45.4 & 35.6 & 39.0 & 46.3 & 77.7 & 52.7 \\
\hline
ViT-B/16 & \bvd{V-55M} & 10M & 54.6 & 71.8 & 51.7 & 36.0 & 40.0 & 48.6 & 79.0 & 55.1 \\
ViT-B/16 & \bvd{V-55M} & 50M & 55.9 & 75.3 & 52.4 & 37.7 & 40.8 & 49.4 & 81.3 & 56.8 \\
\hline
ViT-L/14 & \bvd{V-55M} & 10M & \textbf{63.3} & 78.5 & \textbf{60.6} & \textbf{43.6} & \textbf{44.3} & \textbf{53.7} & 83.4 & 61.9 \\

ViT-L/14 & \bvd{V-55M} & 50M & \textbf{63.3} & \textbf{80.5} & 60.4 & 42.7 & 43.0 & 53.6 & \textbf{84.3} & \textbf{62.0} \\

\hline
\end{tabular}
}
    \floatbottomspace
\end{table}

\begin{table}[t]
\centering
\small
\caption{\textbf{Improved ViCLIP ViT-L/14 results with \bvd{V-*} data using checkpoint merging on \textcolor{bvdred}{classification} and \textcolor{bvdviolet}{retrieval} tasks}. Applying WiSE-FT checkpoint merging~\citep{wortsman2022robust} improves performance of our models. For InternVid-10M-FLT, it narrows the gap to the results reported by~\citep{wang2023internvid}. Compute-matched \bvd{V-10M} still outperforms InternVid-10M-FLT by $2.1\,$pp.}
\label{tab:exp_results_wise_ft}
\resizebox{\textwidth}{!}{%
\begin{tabular}{l c c c c c c c c c c}
\hline
Dataset
& Samples
& WiSE-FT
& \textcolor{bvdred}{\textbf{K400}}
& \textcolor{bvdred}{\textbf{UCF-101}}
& \textcolor{bvdred}{\textbf{HMDB51}}
& \multicolumn{2}{c}{\textcolor{bvdviolet}{\textbf{MSR-VTT}}}
& \multicolumn{2}{c}{\textcolor{bvdviolet}{\textbf{MSVD}}}
& \textbf{Avg} \\
& seen
&
& top-1 & top-1 & top-1
& T2V & V2T
& T2V & V2T
& \\
\hline
InternVid-10M-FLT*
& 10M
& \ding{51}
& 64.8 & {-} & {-}
& 42.4 & 41.3
& 49.1 & 75.1
& {-} \\

InternVid-10M-FLT
& 10M
& \ding{55}
& 61.8 & 78.6 & 55.1
& 38.7 & 39.8
& 50.9 & 74.1
& 58.0 \\

\rowcolor{gray!10}
InternVid-10M-FLT
& 10M
& \ding{51}
& \textbf{65.0} & \textbf{80.8} & 57.3
& 43.5 & 41.7
& 51.4 & 74.5
& 60.2 \\
\hline

\bvd{V-10M}
& 10M
& \ding{55}
& 62.7 & 79.5 & 60.4
& 42.7 & 42.9
& 53.2 & 81.0
& 61.3 \\

\rowcolor{gray!10}
\bvd{V-10M}
& 10M
& \ding{51}
& 64.1 & 80.5 & 60.0
& 44.5 & \textbf{45.3}
& 54.5 & 80.8
& 62.3 \\
\hline

\bvd{V-50M}
& 50M
& \ding{55}
& 63.3 & 79.7 & \textbf{61.4}
& 42.8 & 43.2
& 53.7 & 83.9
& 62.0 \\

\rowcolor{gray!10}
\bvd{V-50M}
& 50M
& \ding{51}
& 64.3 & 79.9 & 61.0
& \textbf{44.6} & 43.7
& \textbf{54.5} & \textbf{84.6}
& \textbf{62.6} \\
\hline
\multicolumn{11}{r}{\footnotesize * results reported in \citep{wang2023internvid}; all other results are evaluated using our pipeline.} \\
\end{tabular}
}
    \floatbottomspace
\end{table}
\paragraph{\viclip{} trained on \laionbvd{} outperforms InternVid-trained models} \Cref{tab:exp_results_viclip_scaling} illustrates performance as a function of increasing dataset size and number of training samples. \bvd{V-10M} and \bvd{V-50M} achieve better performance than InternVid-10M and InternVid-50M. In particular, at 10M samples seen, \bvd{V-10M} and \bvd{V-50M} outperform the more heavily filtered subset InternVid-10M-FLT by 3.3 and 3.5 percentage points respectively, in terms of the overall average metric. %
Furthermore, we observe clear scaling trends with both increasing dataset size and compute, where compute is measured by the number of samples seen; this is particularly notable given the minimal filtering applied to \laionbvd{}.
This trend is further supported by \cref{tab:exp_results_viclip_ref}, where we observe a strictly monotonic increase in average performance with increasing model size and compute. Together, these scaling behaviors and competitive results provide strong evidence that \laionbvd{} is a high-quality dataset for video pre-training.

\paragraph{Improved \viclip{} ViT-L/14 results with WiSE-FT checkpoint merging}
In \cref{tab:exp_results_viclip_scaling}, evaluating the raw InternVid-10M-FLT checkpoint yields lower performance than originally reported in~\cite{wang2023internvid}. As clarified in the official \href{https://github.com/OpenGVLab/InternVideo/issues/139\#issuecomment-2190528342}{InternVideo repository}, the reported numbers in~\cite{wang2023internvid} were obtained using WiSE-FT weight-space ensembling~\cite{wortsman2022robust}, which averages the fine-tuned video checkpoint with the initial \clip{} weights at evaluation time.

To ensure a direct comparison and evaluate the effect of weight ensembling on our models, we apply WiSE-FT to both InternVid-10M-FLT and our \bvd{V-10M}- and \bvd{V-50M}-trained checkpoints (\cref{tab:exp_results_wise_ft}). For InternVid-10M-FLT, we use an interpolation coefficient of $\alpha=0.5$, matching the equal-weight averaging specified by the authors. For our models, we select $\alpha \in \{0.5, 0.6, 0.7, 0.8, 0.9\}$ based on the highest validation average and report test-set performance. Applying WiSE-FT raises the overall average of InternVid-10M-FLT from 58.0 to 60.2, largely accounting for the gap to the originally reported results~\cite{wang2023internvid}. Weight ensembling similarly benefits our models, improving the overall average from 61.3 to 62.3 for \bvd{V-10M} and from 62.0 to 62.6 for \bvd{V-50M}, maintaining a clear margin over InternVid-10M-FLT.

\begin{takeawaybox}
    In a data- and compute-matched comparison, \viclip{} trained on the minimally curated \bvd{V-10M} outperforms the heavily filtered InternVid-10M-FLT baseline by $2.1$ percentage points on average across zero-shot classification and retrieval benchmarks. This indicates that \laionbvd{} is a valuable source for video-text pre-training.
\end{takeawaybox}

\subsection{Audio-text data validation with \clapmodel{}}
\postsubsectionspace
For \clapmodel{} training, we use the captioned audio of \bvd{V-55M} clips between 2 and 30 seconds long (\cref{sec:curation}), in the form of two randomly sampled subsets of 1.7M and 10M clips, \bvd{A-1.7M} and \bvd{A-10M} (\cref{fig:data_curation}).
We evaluate audio performance on one classification benchmark (UrbanSound8K~\citep{Salamon:UrbanSound:ACMMM:14}) and text-audio retrieval~\citep{oncescu2021audio,koepke2022audio} on AudioCaps~\citep{kim2019audiocaps} and Clotho~\citep{drossos2020clotho}, reporting audio-to-text and text-to-audio recall@5. As a summary metric, we report the unweighted mean of all five metrics (UrbanSound8K accuracy and the four retrieval recalls).

\begin{table}[b]
\centering
\small
\caption{\textbf{Average zero-shot audio performance for pure training data sources at 30M samples seen with CLAP models}. Each cell reports the best average score per model scale, aggregated over UrbanSound8K~\citep{Salamon:UrbanSound:ACMMM:14} classification and AudioCaps/Clotho retrieval~\citep{kim2019audiocaps,drossos2020clotho,oncescu2021audio,koepke2022audio}.
Training on \bvd{A-*} alone matches or exceeds LAION-Audio (LA) across model scales, although the strongest results still come from adding curated audio data from AudioSet (LA+AS). Despite minimal curation, \bvd{A-10M} exhibits consistent scaling behavior with increasing model size.}
\label{tab:exp_results_clap_pure}
\resizebox{\textwidth}{!}{%
\begin{tabular}{l r r r r r r r r}
\toprule
Model & Params & Samples & LA & LA+AS & \bvd{A-1.7M} & \bvd{A-10M} \\
\midrule
HTSAT\mbox{-}tiny\mbox{-}Roberta\mbox{-}base & 158M & 30M & 42.1 & \textbf{56.4} & 42.4 & 42.9 \\
HTSAT\mbox{-}base\mbox{-}Roberta\mbox{-}base & 200M & 30M & 43.2 & \textbf{55.5} & 45.7 & 44.8 \\
HTSAT\mbox{-}tiny\mbox{-}Roberta\mbox{-}large & 388M & 30M & 45.0 & \textbf{55.8} & 45.6 & 45.9 \\
HTSAT\mbox{-}base\mbox{-}Roberta\mbox{-}large & 431M & 30M & 44.8 & \textbf{56.6} & 44.5 & 46.8 \\
\bottomrule
\end{tabular}
}
    \floatbottomspace
\end{table}

\begin{table}[t]
\centering
\caption{\textbf{Average zero-shot audio performance with CLAP models for mixed training datasets at $\sim$28M samples seen}.
Each cell reports the best average score per model scale, aggregated over UrbanSound8K~\citep{Salamon:UrbanSound:ACMMM:14} classification and AudioCaps/Clotho retrieval~\citep{kim2019audiocaps, drossos2020clotho}. Across all model scales, replacing LAION-Audio (LA)~\citep{laionclap2023} with \bvd{A-1.7M} yields competitive performance, while incorporating AudioSet (AS) consistently achieves the strongest results.}
\label{tab:exp_results_clap_combined}
\resizebox{\textwidth}{!}{%
\begin{tabular}{l r r r r r r r}
\toprule
Model & Params & Samples & AC+CL+\bvd{A-1.7M} & AC+CL+LA & AC+CL+LA+AS \\
\midrule
HTSAT\mbox{-}tiny\mbox{-}Roberta\mbox{-}base & 158M & 28.2M & 60.6 & 62.2 & \textbf{66.4} \\
HTSAT\mbox{-}base\mbox{-}Roberta\mbox{-}base & 200M & 27.1M & 62.0 & \textbf{64.1} & \textbf{64.1} \\
HTSAT\mbox{-}tiny\mbox{-}Roberta\mbox{-}large & 388M & 27.1M & 61.3 & 63.9 & \textbf{66.6} \\
HTSAT\mbox{-}base\mbox{-}Roberta\mbox{-}large & 431M & 28.2M & 62.3 & 63.7 & \textbf{65.7} \\
\bottomrule
\end{tabular}
}
    \floatbottomspace
\end{table}

\begin{table}[t]
\centering
\caption{\textbf{Scaling behavior of pure \bvd{A-10M} CLAP training}.
We compare best-checkpoint results at roughly 30M and 110M samples seen across four model scales. More seen audio-text samples help primarily at larger model sizes: the 431M model reaches the strongest overall average (48.7) and best retrieval scores at 110M samples seen.}
\label{tab:exp_results_clap_10m_scales}
\resizebox{\textwidth}{!}{%
\begin{tabular}{l l r r r r r r r r r r}
\toprule
 &  &  & & & \multicolumn{2}{c}{\textbf{AudioCaps} R@5} & \multicolumn{2}{c}{\textbf{Clotho} R@5} & \\
\cmidrule(lr){6-7} \cmidrule(lr){8-9}
\textbf{Scale} & \textbf{Model} & \textbf{Params} & \textbf{Samples} & \textbf{US8K} & T2A & A2T & T2A & A2T & \textbf{Avg} \\
\midrule
\raise.17ex\hbox{$\scriptstyle\sim$}30M & HTSAT\mbox{-}tiny\mbox{-}Roberta\mbox{-}base & 158M & 30.0M & 59.2 & 45.3 & 52.5 & 25.1 & 32.2 & \textbf{42.9} \\
 & HTSAT\mbox{-}base\mbox{-}Roberta\mbox{-}base & 200M & 30.0M & 59.0 & 46.9 & 56.5 & 26.3 & 35.5 & \textbf{44.8} \\
 & HTSAT\mbox{-}tiny\mbox{-}Roberta\mbox{-}large & 388M & 30.0M & 60.5 & 48.5 & 58.3 & 26.8 & 35.4 & \textbf{45.9} \\
 & HTSAT\mbox{-}base\mbox{-}Roberta\mbox{-}large & 431M & 30.0M & 64.8 & 48.6 & 59.2 & 26.9 & 34.5 & \textbf{46.8} \\
\midrule
\raise.17ex\hbox{$\scriptstyle\sim$}110M & HTSAT\mbox{-}tiny\mbox{-}Roberta\mbox{-}base & 158M & 110M & 50.9 & 43.5 & 52.2 & 22.5 & 30.9 & \textbf{40.0} \\
 & HTSAT\mbox{-}base\mbox{-}Roberta\mbox{-}base & 200M & 110M & 60.5 & 41.9 & 50.5 & 22.8 & 30.7 & \textbf{41.3} \\
 & HTSAT\mbox{-}tiny\mbox{-}Roberta\mbox{-}large & 388M & 110M & 68.8 & 46.7 & 56.9 & 26.3 & 33.4 & \textbf{46.4} \\
 & HTSAT\mbox{-}base\mbox{-}Roberta\mbox{-}large & 431M & 110M & 70.6 & 50.0 & 59.3 & 27.2 & 36.4 & \textbf{48.7} \\
\bottomrule
\end{tabular}
}
    \floatbottomspace
\end{table}

\paragraph{\bvd{A-10M} outperforms the larger LAION-Audio dataset in single-dataset training} \Cref{tab:exp_results_clap_pure} isolates the effect of the training data source on four model scales, comparing LAION-Audio (LA), LAION-Audio + AudioSet (LA+AS), as well as our \bvd{A-1.7M} and \bvd{A-10M} subsets. Here, we can see that both \bvd{A-1.7M} and \bvd{A-10M} match or exceed LAION-Audio across different model scales for matched samples seen. Only LAION-Audio + AudioSet (which contains data specifically curated for audio tasks) yields better results.
Furthermore, despite minimal curation, both audio subsets exhibit consistent scaling behavior with increasing model size. 

\paragraph{\bvd{A-1.7M} remains competitive with LAION-Audio in mixed-dataset training} \Cref{tab:exp_results_clap_combined} compares \clapmodel{} training on four model scales across three training sets: AudioCaps + Clotho + \bvd{A-1.7M} (AC+CL+\bvd{A-1.7M}); AudioCaps + Clotho + LAION-Audio (AC+CL+LA); and AudioCaps + Clotho + LAION-Audio + AudioSet (AC+CL+LA+AS), which matches the original LAION-CLAP~\citep{laionclap2023} training data.
We observe that despite only minimal curation of \bvd{A-1.7M}, AC+CL+\bvd{A-1.7M} remains close to AC+CL+LA while AC+CL+LA+AS, containing the highly curated AudioSet, is strongest overall.

\paragraph{\bvd{A-10M} exhibits consistent scaling behavior across model sizes}
\Cref{tab:exp_results_clap_10m_scales} shows the scaling behavior of \bvd{A-10M} at two compute scales (30M and 110M samples seen) and four model sizes. Increasing model size consistently improves average benchmark performance. Increasing the number of samples helps only for larger models (388M+). This is an expected bottleneck effect, where smaller model scales cannot benefit from increasing sample scale, with a further factor being the diminishing gains from the high number of repetitions (110M samples scale) over only 10M unique samples in \bvd{A-10M}. The scaling effect is  also present in mixed dataset training (\Cref{tab:exp_results_clap_combined}). However, it is weaker, as the highly curated small scale AudioCaps/Clotho training datasets dominate the model performance.

\begin{takeawaybox}
    \laionbvd{} provides a viable source of audio data for multimodal pre-training, supported by consistent performance across single- and mixed-dataset settings and favorable scaling behavior.
\end{takeawaybox}

\subsection{Image-text validation with CLIP}\label{sec:exp_results}
\postsubsectionspace

\begin{table}[t]
\centering
\caption{\textbf{Image-text training with CLIP across datasets and scales}.
        \bvd{I-300M} achieves the strongest COCO retrieval results, while web-image datasets remain stronger on ImageNet-style classification.}
\resizebox{\textwidth}{!}{%
        \sisetup{
            detect-weight=true,
        }
        \renewcommand{\arraystretch}{1.4}
        \settowidth{\unitwidth}{M}
        \newcolumntype{U}{>{\columncolor{white}[0pt][\tabcolsep]\raggedright\arraybackslash}p{\dimexpr\unitwidth}}
        \newlength\sampleswidth
        \settowidth{\sampleswidth}{\textbf{Samples}}
        \newcolumntype{N}{>{\raggedleft\arraybackslash}p{\dimexpr\sampleswidth-\unitwidth}}
        \begin{tabular}{l l r S[table-format=1.2] S[table-format=1.2] S[table-format=1.2] S[table-format=1.2] S[table-format=1.2] S[table-format=1.2]}
        \toprule
        & & & Acc &  \multicolumn{2}{c}{\textbf{COCO Retrieval} R@5} & \multicolumn{3}{c}{Acc} \\
        \cmidrule(lr){4-4} \cmidrule(lr){5-6} \cmidrule(lr){7-9}
        \textbf{Model} & \textbf{Dataset} & \multicolumn{1}{c}{\textbf{Samples}} & {\textbf{ImNet-1k}} & \textbf{T2I} & \textbf{I2T} & \textbf{ImNet-R} & \textbf{ImNet-Sketch} & \textbf{ImNet-V2} \\
\midrule
ViT-B/32 & Re-LAION & 30.7M & 0.17 & 0.20 & 0.32 & 0.22 & 0.10 & 0.15 \\
 &  & 64M & 0.26 & 0.30 & 0.45 & 0.32 & 0.17 & 0.22 \\
 &  & 128M & 0.35 & 0.38 & 0.53 & 0.40 & 0.24 & 0.28 \\
 &  & 300M & 0.44 & 0.46 & 0.64 & 0.52 & 0.32 & 0.37 \\
ViT-B/32 & DataComp & 30.7M & 0.20 & 0.21 & 0.32 & 0.25 & 0.12 & 0.17 \\
 &  & 64M & 0.31 & 0.29 & 0.43 & 0.36 & 0.21 & 0.26 \\
 &  & 128M & 0.40 & 0.37 & 0.54 & 0.45 & 0.29 & 0.33 \\
 &  & 300M & \textbf{0.50} & 0.45 & 0.63 & \textbf{0.57} & \textbf{0.39} & \textbf{0.42} \\
\rowcolor{black!10} ViT-B/32 & \bvd{I-300M} & 30.7M & 0.08 & 0.27 & 0.40 & 0.13 & 0.04 & 0.08 \\
\rowcolor{black!10} &  & 64M & 0.13 & 0.40 & 0.56 & 0.18 & 0.06 & 0.11 \\
\rowcolor{black!10} &  & 128M & 0.19 & 0.48 & 0.67 & 0.26 & 0.11 & 0.16 \\
\rowcolor{black!10} &  & 300M & 0.24 & \textbf{0.56} & \textbf{0.74} & 0.34 & 0.16 & 0.21 \\
\midrule
ViT-B/16 & Re-LAION & 128M & 0.41 & 0.45 & 0.62 & 0.48 & 0.29 & 0.34 \\
 &  & 300M & 0.51 & 0.53 & 0.71 & 0.59 & 0.38 & 0.44 \\
ViT-B/16 & DataComp & 30.7M & 0.22 & 0.26 & 0.38 & 0.28 & 0.15 & 0.21 \\
 &  & 64M & 0.38 & 0.36 & 0.52 & 0.42 & 0.26 & 0.32 \\
 &  & 128M & 0.47 & 0.43 & 0.60 & 0.52 & 0.34 & 0.40 \\
 &  & 300M & \textbf{0.58} & 0.52 & 0.70 & \textbf{0.64} & \textbf{0.44} & \textbf{0.49} \\
\rowcolor{black!10} ViT-B/16 & \bvd{I-300M} & 30.7M & 0.10 & 0.35 & 0.51 & 0.16 & 0.05 & 0.10 \\
\rowcolor{black!10} &  & 64M & 0.18 & 0.50 & 0.67 & 0.24 & 0.09 & 0.15 \\
\rowcolor{black!10} &  & 128M & 0.22 & 0.56 & 0.74 & 0.30 & 0.14 & 0.19 \\
\rowcolor{black!10} &  & 300M & 0.28 & \textbf{0.63} & \textbf{0.80} & 0.40 & 0.19 & 0.23 \\
\bottomrule
\end{tabular}
}
\label{tab:exp_results_clip}
    \floatbottomspace
\end{table}

We train CLIP models on extracted frame-caption pairs from \bvd{I-300M} and compare them to models trained on DataComp-1B~\citep{gadre2023datacomp} and Re-LAION-2B~\citep{relaion} using the same training recipe and data scales, evaluating on zero-shot ImageNet-1k classification and MS-COCO retrieval (\cref{tab:exp_results_clip}).

\paragraph{\bvd{I-300M} achieves strong retrieval performance} Overall, models trained on \bvd{I-300M} perform strongly on text-to-image and image-to-text retrieval, while lagging behind on ImageNet-1k zero-shot classification (see \cref{app:caption_analysis} for further analysis).
We observe comparable performances for CLIP models trained on \bvd{I-300M} and models trained on DataComp-1B recaptioned with our captioning pipeline (e.g.\ 0.23 ImageNet-1k accuracy compared to 0.19 with \bvd{I-300M} when training on 128M frames as can be seen in \cref{tab:caption-validation,tab:exp_results_clip}).

In addition to ImageNet-1k, we also evaluate our models on ImageNet-R~\citep{hendrycks2021many}, ImageNet-Sketch~\citep{wang2019learning}, and ImageNet-V2~\citep{recht2019imagenet}, which further support the generalization capabilities of models trained on \bvd{I-300M}.

\begin{figure}
    \centering
    \subfloat[ImageNet-1k zero-shot classification error rate]
    {\includegraphics[width=0.49\textwidth]{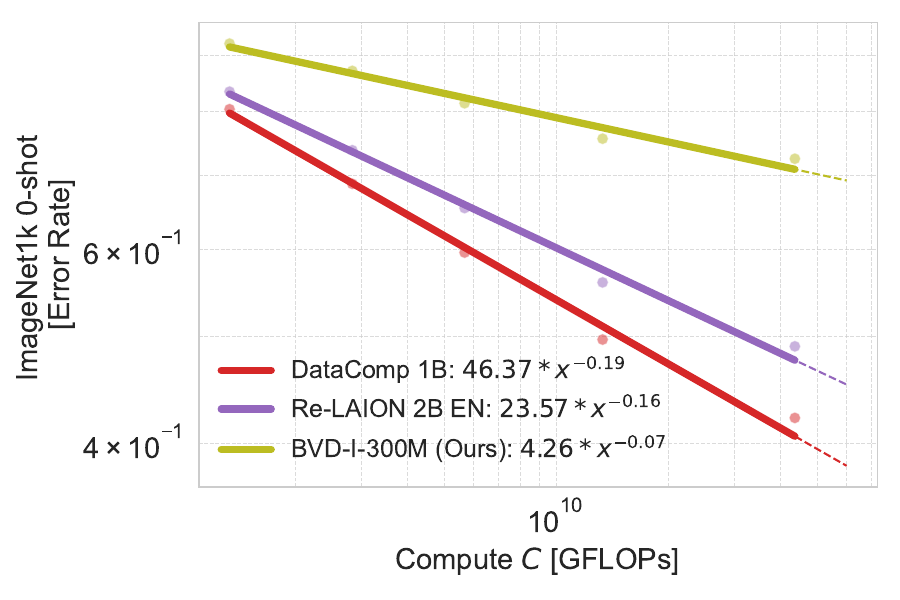}}
     \hfill
    \subfloat[MS-COCO image retrieval R@5 error rate]
    {\includegraphics[width=0.49\textwidth]{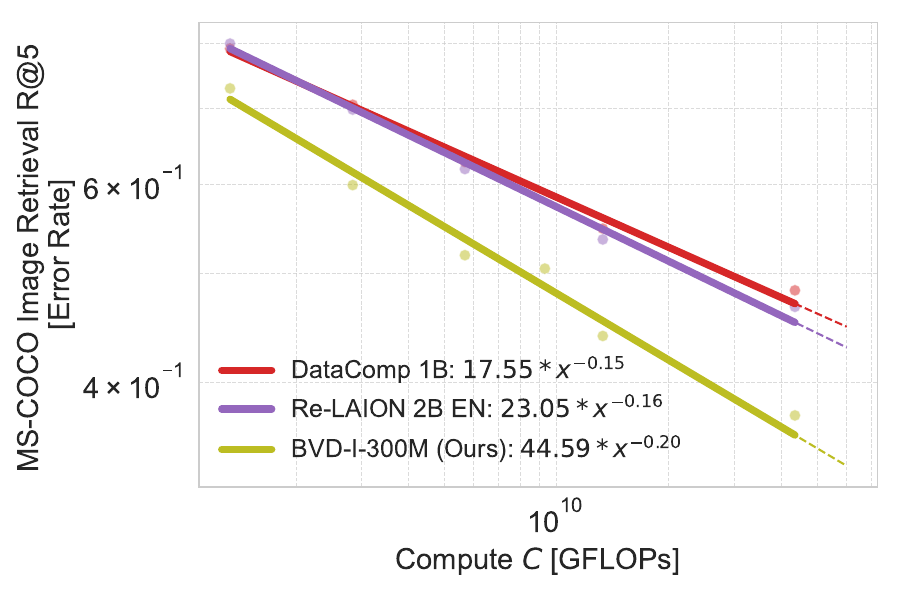}}
    \caption{
        \textbf{Image-text scaling trends for \bvd{I-300M}, Re-LAION, and DataComp-1B}.
        \bvd{I-300M} scales especially well on retrieval, while its ImageNet zero-shot classification performance remains weaker than web alt-text baselines.
    }
    \label{fig:scaling}
    \floatbottomspace
\end{figure}

\paragraph{\bvd{I-300M} shows strong scaling trends for retrieval tasks} The plots in \cref{fig:scaling} show strong scaling trends for \bvd{I-300M} across tasks. While all datasets yield improved performance when using larger sets of image-text pairs for training, \bvd{I-300M} exhibits weaker scaling for ImageNet-1k zero-shot classification compared to Re-LAION and DataComp. This again confirms that its synthetic captions may be less effective for fine-grained, label-centric classification. However, we emphasize that strong retrieval performance is indicative of high-quality embeddings, making \bvd{I-300M} particularly well-suited for representation learning and downstream applications such as similarity search or kNN-based indexing. In this context, the lower ImageNet-1k performance is less concerning and likely reflects a weaker alignment with the specific class distribution and biases of ImageNet-1k: biases that are explicitly or implicitly encoded in datasets like DataComp (e.g., via clustering towards ImageNet classes) and Re-LAION/OpenCLIP (e.g., via CLIP-based filtering grounded in WordNet hierarchies). In contrast, \bvd{I-300M} demonstrates strong scaling behavior on MS-COCO text-to-image retrieval, outperforming other datasets with a steeper slope and lower error rates when using more data, further underscoring its effectiveness for retrieval-focused applications.

\begin{takeawaybox}
Frame-caption pairs from \bvd{I-300M} provide strong supervision for image-text retrieval and scale favorably with data size, but yield weaker performance on classification benchmarks due to differences in caption style and class coverage. This shows that videos from \laionbvd{} can be used as a new source of image-text pairs.
\end{takeawaybox}

\presectionspace
\section{Limitations}
\postsectionspace
\paragraph{Caption quality and style}
All captions in \laionbvd{} are automatically generated by small captioning models and are intentionally short. This enables scalable processing across large volumes of video data, but it constrains the richness of the captions. It may also introduce systematic errors or biases from the captioning model.

\paragraph{Model scope}
Our evaluation focuses on contrastive video-text training using \viclip{} and does not cover generative paradigms such as video-language models or diffusion-based approaches. While we believe that the validation experiments with \viclip{}, \clapmodel{}, and CLIP provide a useful signal of dataset quality, future work should further evaluate the data on generative tasks.

\paragraph{Audio-visual integration}
We evaluate video-text, audio-text, and image-text settings separately, and do not train joint audio-visual models. This leaves open how well \laionbvd{} supports learning unified representations across modalities, particularly for tasks that require tight synchronization between visual and auditory signals.

\paragraph{Data biases}
\laionbvd{} is constructed from video platforms with minimal filtering. While the underlying platforms apply moderation, the dataset may still contain biases, stereotypes, or uneven representation across languages, regions, and topics. Models trained on this dataset can inherit such biases.

\presectionspace
\section{Discussion and Conclusion}
\label{sec:discussion_conclusion}
\postsectionspace

The \laionbvd{} dataset is designed for multimodal pre-training at scale. It features 1.3B video links (\bvd{URLs}), 13.5 million video hours (\bvd{RAW}), as well as 55M video, 10M audio, and 300M frame captions (\bvd{V-55M}, \bvd{A-10M}, \bvd{I-300M}).

Our analysis shows that all modalities (video, audio, individual frames) provide a strong training signal. \viclip{} models trained on \bvd{V-10M} outperform InternVid-based models on matched model and data scale. For audio, \clapmodel{} trained on \bvd{A-*} achieves competitive results despite minimal curation, and CLIP models trained on extracted video frames (\bvd{I-300M}) perform strongly on retrieval benchmarks. Across modalities, we observe consistent scaling trends with increasing data and compute, indicating that \laionbvd{} is a reliable source for multimodal pre-training.

\laionbvd{} broadens access to large-scale open web video and supports more reproducible multimodal foundation model research. By releasing the dataset and metadata to research institutions, and by documenting a scalable curation pipeline, we aim to lower the barrier to studying video, audio, and image data jointly at this scale.

\section*{Acknowledgments}
This work was supported in part by the German Federal Ministry of Research, Technology and Space (BMFTR) under grants FKZ 01IS24060, FKZ 01IS18039A, FKZ 16IS24079A (XEI), 01IS24085C (OPENHAFM), 01IS22094B (WestAI – AI Service Center West), and 16HPC117K (MINERVA), as well as by the German Research Foundation (DFG) through SFB 1233 (project number 276693517) and utilized compute resources at the Tübingen Machine Learning Cloud, DFG FKZ INST 37/1057-1 FUGG. AH acknowledges support from the International Max Planck Research School for Intelligent Systems (IMPRS-IS). MB additionally acknowledges support from Coefficient Giving funded by the Good Science Ventures Foundation (Open Philanthropy). MB, MN, MC, and JJ further acknowledge funding from the European Union Horizon Programme under grant no. 101214398 (ELLIOT), co-funding from the Digital Europe Programme under grant no. 101195233 (openEuroLLM) and grant no. 101198470 (LLMs4EU), and support from the EuroHPC Joint Undertaking under grant no. 101182737 (MINERVA). WB acknowledges financial support via an Emmy Noether Grant funded by the German Research Foundation (DFG) under grant no. BR 6382/1-1 and WB is a member of the Machine Learning Cluster of Excellence, EXC number 2064/1 - Project number 390727645.

We gratefully acknowledge the  Gauss Centre for Supercomputing e.V. (GCS) for providing computing resources through the John von Neumann Institute for Computing (NIC) on the JUWELS Booster and JUPITER supercomputers at the  Jülich Supercomputing Centre (JSC). We further acknowledge the EuroHPC Joint Undertaking for computing time and storage on the LEONARDO supercomputer hosted by CINECA (Italy) and the LEONARDO consortium through the EuroHPC Extreme Access grant EHPC-EXT-2023E02-068. We also acknowledge storage resources on JUST operated by JSC and supported by the Helmholtz Data Federation (HDF), computing time granted by JARA and JSC on the JURECA supercomputer, access to the JEDI prototype system through the JUREAP (JUPITER Early Access Program), and computing time provided through the Gauss AI Competition (reformo) on JUPITER via GCS and BMFTR.

LAION further acknowledges a public storage grant from  Hugging Face, enabling broad community access to open-source research outputs through Hugging Face repositories. We also thank the teams operating the supercomputing facilities for their support, especially Björn Hagemeier for assistance with data storage and handling.

\bibliography{main}

\newpage
\appendix
\section{Additional Experimental Details}\label{sec:experimental_details}

\subsection{Video-Language (\viclip{})}

In this section, we present additional details about our video-language experiments. 

\subsubsection{Experimental setup}

ViCLIP~\citep{wang2023internvid} models consist of two components, a video encoder and a text encoder. The video and text encoders are initialized from a pre-trained CLIP model's vision and text encoder respectively, with a modification of the vision encoder to process a set of frames instead of a single image, while the text encoder remains unchanged. The main architectural change compared to standard CLIP is that the vision encoder receives patches from all the frames, on which we apply both spatial and temporal positional embeddings. For attention layers, any patch token can attend to any patch token across the frames. Training is performed using standard InfoNCE loss~\citep{oord2018representation} computed using video and text embeddings.

We train our ViCLIP models by initializing the video frame encoder and text encoder from CLIP pre-trained checkpoints on DataComp-1B~\citep{gadre2023datacomp}. We uniformly sample 8 frames from each video following~\citep{wang2023internvid}.
For every training run in \cref{tab:exp_results_viclip_scaling,tab:exp_results_viclip_ref}, we performed learning rate hyperparameter sweeps covering \texttt{1e-6}, \texttt{4e-6}, \texttt{1e-5}, \texttt{4e-5}, \texttt{1e-4}, \texttt{4e-4}.
We also experimented with two global batch sizes of \texttt{8k} and \texttt{32k}.
For optimization, we used AdamW~\citep{adamw} with $\beta_1 = 0.9$ and $\beta_2 = 0.98$, a weight decay of $0.2$, and gradient clipping with a norm of $1$. We used the standard cosine learning rate schedule with a warmup of 100 in all experiments. We trained our models on 100 NVIDIA GH200 GPUs. To construct a validation set, we randomly sampled a maximum of 10K examples from the training sets of each downstream dataset (K400, UCF-101, HMDB51, MSR-VTT, MSVD) and evaluated all our trained models on the validation sets. We then used the average metric (defined in \cref{sec:viclip_results}) on the validation sets for hyperparameter selection. In \cref{tab:viclip_hyperparameters}, we show the best hyperparameters used to train our ViCLIP models and report the final test metrics of each task and the overall average in \cref{tab:exp_results_viclip_scaling,tab:exp_results_viclip_ref}.

\begin{table}[h]
\centering
\caption{\textbf{Hyperparameters used for training ViCLIP models}.}
\label{tab:viclip_hyperparameters}
\begin{tabular}{lccccccc}
\hline
Model & Dataset & Samples seen & Batch Size & LR & WD & FLOPs & GPU-hours \\
\hline
ViT-B/32 & \bvd{V-55M} & 10M & 32k & 1e-4 & 0.2 & 2.22e18 & 12 \\
ViT-B/32 & \bvd{V-55M} & 50M & 32k & 1e-4 & 0.2 & 1.11e19 & 65 \\
ViT-B/16 & \bvd{V-55M} & 10M & 32k & 1e-4 & 0.2 & 8.1e18 & 25 \\
ViT-B/16 & \bvd{V-55M} & 50M & 32k & 4e-5 & 0.2 & 4.2e19 & 125 \\
ViT-L/14 & \bvd{V-10M} & 10M & 32k & 1e-4 & 0.2 & 3.9e19 & 90 \\
ViT-L/14 & \bvd{V-10M} & 50M & 32k & 4e-5 & 0.2 & 1.89e20 & 455 \\
ViT-L/14 & \bvd{V-50M} & 10M & 32k & 1e-4 & 0.2 & 3.9e19 & 90 \\
ViT-L/14 & \bvd{V-55M} & 10M & 32k & 1e-4 & 0.2 & 3.9e19 & 90 \\
ViT-L/14 & \bvd{V-50M} & 50M & 32k & 4e-5 & 0.2  & 1.89e20 & 455 \\
ViT-L/14 & \bvd{V-55M} & 50M & 32k & 4e-5 & 0.2 & 1.89e20 & 455 \\
\hline
\end{tabular}
\end{table}

\subsubsection{Additional results}

\paragraph{Statistical uncertainty analysis}
To assess whether observed differences exceed run-to-run variation, \cref{tab:bvd_10m_vs_50m} reports mean, standard deviation, and 95\% confidence intervals across runs for \bvd{V-10M} and \bvd{V-50M} at fixed training compute (50M samples seen). While several metrics show only minor differences within the uncertainty range, we observe statistically clear improvements on HMDB51, MSR-VTT V2T, MSVD T2V, and the overall average metric when training on the larger dataset.

\begin{table}[h]
\centering
\small
\caption{
\textbf{Comparison between \bvd{V-10M} and \bvd{V-50M} at fixed training compute (50M samples seen).} 
We report mean and standard deviation across five runs with different seeds together with the performance difference and corresponding 95\% confidence interval. Larger datasets primarily improve retrieval performance and HMDB51 classification, while gains on K400 and UCF-101 remain within the estimated uncertainty range.
}
\label{tab:bvd_10m_vs_50m}
\resizebox{\textwidth}{!}{%
\begin{tabular}{l l c c c c c}
\hline
Benchmark & Metric & \bvd{V-10M} & \bvd{V-50M} & Difference & 95\% CI & Conclusion \\
\hline
K400 & top-1 & 63.78$_{\pm0.15}$ & 63.84$_{\pm0.17}$ & +0.06 & [-0.20, +0.32] & not clear \\
UCF-101 & top-1 & 80.28$_{\pm0.37}$ & 80.42$_{\pm0.25}$ & +0.14 & [-0.35, +0.63] & not clear \\
HMDB51 & top-1 & 55.50$_{\pm0.23}$ & 57.48$_{\pm0.22}$ & \textbf{+1.98} & [+1.54, +2.42] & clear improvement \\
MSR-VTT & T2V & 42.58$_{\pm0.39}$ & 42.62$_{\pm0.20}$ & +0.04 & [-0.49, +0.57] & not clear \\
 & V2T & 42.32$_{\pm0.26}$ & 43.28$_{\pm0.54}$ & \textbf{+0.96} & [+0.06, +1.86] & clear improvement \\
MSVD & T2V & 53.30$_{\pm0.07}$ & 53.52$_{\pm0.11}$ & \textbf{+0.22} & [+0.06, +0.38] & clear improvement \\
 & V2T & 83.32$_{\pm0.46}$ & 83.76$_{\pm0.79}$ & +0.44 & [-0.78, +1.66] & not clear \\
Overall & Avg & 60.94$_{\pm0.11}$ & 61.52$_{\pm0.18}$ & \textbf{+0.58} & [+0.40, +0.76] & clear improvement \\
\hline
\end{tabular}
}
\vspace{-0.4cm}
\end{table}

\paragraph{Overlap and decontamination analysis.}
We perform an overlap analysis between \bvd{V-55M} and the K400, MSVD, and MSR-VTT test sets using unique YouTube video IDs which is shown in \cref{tab:dataset_overlap}. We then construct decontaminated versions of the downstream test sets by removing all examples whose YouTube video IDs are present in \bvd{V-55M}. The resulting test-set sizes are shown in \cref{tab:decontamination_sizes}. For MSR-VTT and MSVD, the 885 and 522 entries in \cref{tab:dataset_overlap} denote the number of unique YouTube video IDs, whereas the corresponding test sets contain 1,000 and 670 examples, respectively. This discrepancy arises because multiple test examples may correspond to clips from the same source video. We re-evaluate our models on the decontaminated MSR-VTT and MSVD test sets and compare the results with those obtained on the original test sets in \cref{tab:msrvtt_decontamination,tab:msvd_decontamination} respectively. Overall, performance on the decontaminated test sets remains comparable to that on the original test sets, indicating that the observed overlap with \bvd{V-55M} does not materially affect the reported results.

\begin{table}[tp]
    \centering
    \caption{\textbf{Overlap of \bvd{V-55M} with evaluation datasets.} We report the number and percentage of unique YouTube video IDs from K400, MSR-VTT, and MSVD that overlap with \bvd{V-55M}.}
    \label{tab:dataset_overlap}
    \begin{tabular}{lrrr}
        \toprule
        \textbf{Dataset} & \textbf{Overlap} & \textbf{Unique Videos} & \textbf{Overlap (\%)} \\
        \midrule
        K400    & 103 & 39,805 & 0.26 \\
        MSR-VTT & 49  & 885    & 5.5  \\
        MSVD    & 33  & 522    & 6.3  \\
        \bottomrule
    \end{tabular}
\end{table}

\begin{table}[tp]
    \centering
    \caption{\textbf{Evaluation dataset sizes after decontamination.} We report the original and decontaminated test-set sizes after removing examples whose YouTube video IDs are present in \bvd{V-55M}, together with the number of removed examples ($\Delta$).}
    \label{tab:decontamination_sizes}
    \begin{tabular}{lrrr}
        \toprule
        \textbf{Dataset} & \textbf{Original} & \textbf{Decontaminated} & $\Delta$ \\
        \midrule
        K400    & 38,686 & 38,583 & 103 \\
        MSVD    & 670    & 631    & 39  \\
        MSR-VTT & 1,000  & 943    & 57  \\
        \bottomrule
    \end{tabular}
\end{table}

\begin{table*}[tp]
    \centering
    \caption{\textbf{Decontamination has minimal impact on MSR-VTT retrieval performance.} We compare text-to-video (T2V) and video-to-text (V2T) retrieval performance on the original and decontaminated MSR-VTT test sets across pre-training datasets, model sizes, and training compute.}
    \label{tab:msrvtt_decontamination}
    \resizebox{\textwidth}{!}{
    \begin{tabular}{llrrrrrr}
        \toprule
        \textbf{Model} &
        \textbf{Pre-training Dataset} &
        \textbf{Samples Seen} &
        \multicolumn{2}{c}{\textbf{T2V}} &
        \multicolumn{2}{c}{\textbf{V2T}} \\
        \cmidrule(lr){4-5}
        \cmidrule(lr){6-7}
        & & &
        \textbf{Original} & \textbf{Decont.} &
        \textbf{Original} & \textbf{Decont.} \\
        \midrule
        ViCLIP ViT-L/14 & \bvd{V-10M} & 10M & 42.7 & 42.3 & 42.9 & 42.9 \\
        ViCLIP ViT-L/14 & \bvd{V-10M} & 50M & 41.3 & 41.0 & 43.1 & 42.9 \\
        ViCLIP ViT-L/14 & \bvd{V-50M} & 10M & 43.5 & 43.3 & 42.6 & 42.7 \\
        ViCLIP ViT-L/14 & \bvd{V-50M} & 50M & 42.8 & 42.5 & 43.2 & 43.4 \\
        \midrule
        ViCLIP ViT-B/32 & \bvd{V-55M} & 10M & 34.2 & 34.5 & 38.1 & 38.5 \\
        ViCLIP ViT-B/32 & \bvd{V-55M} & 50M & 35.6 & 35.7 & 39.0 & 39.4 \\
        ViCLIP ViT-B/16 & \bvd{V-55M} & 10M & 36.0 & 36.1 & 40.0 & 40.2 \\
        ViCLIP ViT-B/16 & \bvd{V-55M} & 50M & 37.7 & 37.8 & 40.8 & 41.1 \\
        ViCLIP ViT-L/14 & \bvd{V-55M} & 10M & 43.6 & 43.6 & 44.3 & 44.0 \\
        ViCLIP ViT-L/14 & \bvd{V-55M} & 50M & 42.2 & 41.8 & 43.8 & 43.7 \\
        \bottomrule
    \end{tabular}
    }
\end{table*}

\begin{table*}[tp]
    \centering
    \caption{\textbf{Decontamination has minimal impact on MSVD retrieval performance.} We compare text-to-video (T2V) and video-to-text (V2T) retrieval performance on the original and decontaminated MSVD test sets across pre-training datasets, model sizes, and training compute.}
    \label{tab:msvd_decontamination}
    \resizebox{\textwidth}{!}{
    \begin{tabular}{llrrrrrr}
        \toprule
        \textbf{Model} &
        \textbf{Pre-training Dataset} &
        \textbf{Samples Seen} &
        \multicolumn{2}{c}{\textbf{T2V}} &
        \multicolumn{2}{c}{\textbf{V2T}} \\
        \cmidrule(lr){4-5}
        \cmidrule(lr){6-7}
        & & &
        \textbf{Original} & \textbf{Decont.} &
        \textbf{Original} & \textbf{Decont.} \\
        \midrule
        ViCLIP ViT-L/14 & \bvd{V-10M} & 10M & 53.2 & 53.1 & 81.0 & 80.8 \\
        ViCLIP ViT-L/14 & \bvd{V-10M} & 50M & 53.2 & 53.1 & 83.0 & 83.0 \\
        ViCLIP ViT-L/14 & \bvd{V-50M} & 10M & 53.6 & 53.5 & 82.1 & 81.7 \\
        ViCLIP ViT-L/14 & \bvd{V-50M} & 50M & 53.7 & 53.5 & 83.9 & 83.8 \\
        \midrule
        ViCLIP ViT-B/32 & \bvd{V-55M} & 10M & 44.2 & 44.0 & 74.8 & 74.7 \\
        ViCLIP ViT-B/32 & \bvd{V-55M} & 50M & 46.3 & 46.3 & 77.7 & 77.9 \\
        ViCLIP ViT-B/16 & \bvd{V-55M} & 10M & 48.6 & 48.6 & 79.0 & 79.0 \\
        ViCLIP ViT-B/16 & \bvd{V-55M} & 50M & 49.4 & 49.3 & 81.3 & 80.9 \\
        ViCLIP ViT-L/14 & \bvd{V-55M} & 10M & 53.7 & 53.6 & 83.4 & 83.0 \\
        ViCLIP ViT-L/14 & \bvd{V-55M} & 50M & 53.6 & 53.5 & 84.0 & 84.0 \\
        \bottomrule
    \end{tabular}
    }
\end{table*}

\paragraph{Human audit of video captions}
To provide a direct assessment of video-caption faithfulness, we conducted a human audit of 134 captioned videos. We classified each caption as: (i) \textit{accurate}, when it faithfully described the video; (ii) \textit{minor error}, when it was broadly correct but contained a localized mistake, such as an incorrect action, object, or color; or (iii) \textit{major error}, when it substantially misrepresented the video. The results are as follows:
\begin{itemize}
    \item Accurate: 106/134 (79.1\%)
    \item Minor error: 25/134 (18.7\%)
    \item Major error: 3/134 (2.2\%)
    \item No major error: 131/134 (97.8\%)
\end{itemize}

The observed errors involved incorrect actions (17 cases), objects (7), or scene descriptions (4). We emphasize that a primary goal of this study is to validate the quality of the dataset itself. The audit indicates that most video captions are faithful, while the downstream improvements over InternVid (\cref{tab:exp_results_viclip_scaling}) suggest that they provide useful training supervision.

\subsection{Audio-Language (\clapmodel{})}

\subsubsection{Experimental setup}

We validate the audio subsets of \laionbvd{} using \clapmodel{}~\citep{laionclap2023}, a contrastive audio-language model trained on paired audio-caption data. Audio waveforms are resampled to \SI{48}{\kilo\hertz} and converted into log Mel spectrogram representations before being processed by the audio encoder. Text captions are tokenized and encoded using the CLAP text encoder.

The audio and text embeddings are projected into a shared embedding space and trained using a symmetric contrastive loss over matched and mismatched audio-text pairs. We train models on \bvd{A-1.7M} and \bvd{A-10M} to study scaling behavior across both dataset and model size. \cref{tab:computational_cost_clap} reports the computational training cost of different \clapmodel{} backbones at a fixed number of training samples seen.

Evaluation follows standard CLAP protocols using zero-shot classification on UrbanSound8K and retrieval benchmarks on AudioCaps and Clotho. We report text-to-audio and audio-to-text Recall@5 metrics.

\begin{table}[htbp]
\centering
\caption{\textbf{Computational training cost across different \clapmodel{} backbones at a fixed number of training samples seen.}}
\label{tab:computational_cost_clap}
\begin{tabular}{l r r r r r r r}
\toprule
Model & Params & Samples & FLOPs per sample & Total FLOPs \\
\midrule
HTSAT\mbox{-}tiny\mbox{-}Roberta\mbox{-}base & 158M & 28.2M & 3.3e11 & 2.8e19 \\
HTSAT\mbox{-}base\mbox{-}Roberta\mbox{-}base & 200M & 27.1M & 3.9e11 & 3.3e19 \\
HTSAT\mbox{-}tiny\mbox{-}Roberta\mbox{-}large & 388M & 27.1M & 1.05e12 & 8.7e19 \\
HTSAT\mbox{-}base\mbox{-}Roberta\mbox{-}large & 431M & 28.2M & 1.11e12 & 9.3e19 \\
\bottomrule
\end{tabular}
\end{table}

\subsubsection{Additional results}
In this section, we present additional results to complement our audio-language validation. \Cref{tab:exp_results_clap_ref} compares our \clapmodel{} pipeline to the original LAION-CLAP model trained on the same dataset. Our implementation consistently achieves higher performance, showing the correctness of our training pipeline.

\begin{table}[h]
\centering
\small
\caption{\textbf{Audio-language comparison at matched 158M model scale using different training datasets}.
Our open CLAP pipeline reproduces and exceeds the published LAION-CLAP~\citep{laionclap2023} baseline at comparable or lower training compute. Replacing LAION-Audio with \bvd{A-1.7M} remains competitive overall and is particularly strong on UrbanSound8K and AudioCaps retrieval.}
\label{tab:exp_results_clap_ref}
\resizebox{\textwidth}{!}{%
\begin{tabular}{l r r r r r r r r r}
\toprule
Trainset & D (K) & S (M) & C (FLOPs) & Avg & US8K & AC aR@5 & AC tR@5 & Cl aR@5 & Cl tR@5 \\
\midrule
LAION\mbox{-}CLAP (AC+CL+LA+AS) & 2621 & 118 & $1.17 \times 10^{20}$ & 61.4 & 73.2 & 70.5 & 79.5 & 39.0 & 44.6 \\
\midrule
AC+CL+LA & 590 & 27.1 & $2.73 \times 10^{19}$ & 62.2 & 78.6 & 69.0 & 74.7 & \textbf{41.3} & 47.5 \\
\midrule
AC+CL+LA+AS & 2621 & 27.0 & $2.70 \times 10^{19}$ & \textbf{66.4} & \textbf{79.6} & \textbf{75.5} & \textbf{85.9} & 41.1 & \textbf{49.7} \\
\midrule
AC+CL+BVD\mbox{-}A\mbox{-}1.7M & 1767 & 28.2 & $2.82 \times 10^{19}$ & 60.6 & 76.2 & 70.3 & 80.4 & 32.1 & 43.8 \\
\bottomrule
\end{tabular}
}
\end{table}

\subsubsection{Human audit of audio captions}
We conducted a small human audit of the audio captions using 134 randomly sampled captioned clips. The results are:
\begin{itemize}
    \item Accurate: 106/134 (79.1\%)
    \item Minor error: 20/134 (14.9\%)
    \item Major error: 8/134 (6.0\%)
    \item No major error: 126/134 (94.0\%)
\end{itemize}

The main hallucination types involved incorrect sound events (10 cases), speech content (5), music descriptions (4), and speaker attributes (1). These results show that most audio captions are faithful, while also revealing that audio captioning remains more prone to substantial errors than video captioning.

\subsubsection{Contamination analysis of Audio Flamingo 3 with downstream test sets}

We cross-reference the released Audio Flamingo 3 (AF3) training manifests with the AudioCaps and Clotho test sets. For Clotho, we find no test-set overlap; the 367 matching clips all belong to the Clotho training split. For AudioCaps, 356 of 975 test video IDs occur in the AF3 AudioSkills subset. However, these examples are paired with AudioSet-derived captions rather than the AudioCaps reference captions used for evaluation. Thus, AF3 was exposed to a subset of AudioCaps test audio, but not to its evaluation captions.

To test whether this overlap affects retrieval performance, we re-evaluate all 140 trained runs (2,118 checkpoints) on an AF3-decontaminated AudioCaps subset. Of the 975 test clips, 619 are absent from AF3 training and 607 are available in our evaluation shards. Since reducing the retrieval gallery from 963 to 607 clips itself increases R@5 by approximately 8.3 percentage points, we compare the 607 clean clips with a fixed randomly sampled control set of the same size.

The resulting decontamination effects are shown in \cref{tab:af3_decontamination}. Models trained on \bvd{A-*} are not more affected by removing AF3-seen examples; pooled \bvd{A-*} models show a smaller reduction than models trained without \bvd{A-*}. Instead, the effect is larger for retrieval models trained directly on AudioCaps or AudioSet-derived data, independent of whether \bvd{A-*} is included. This suggests that AF3 overlap does not explain the retrieval gains observed for \bvd{A-*}.

\paragraph{Caption-style analysis.}
We further compare caption statistics across \bvd{A-1.7M}, AudioCaps, and Clotho (\cref{tab:caption_length_distribution}). AudioCaps and Clotho have similar mean caption lengths (approximately 8 and 11 words), while \bvd{A-1.7M} has a broader length distribution. If AF3 recaptioning transferred benchmark-specific caption style to our audio captions, we would also expect improved performance on Clotho, whose training captions were available to AF3 as metadata. Instead, \bvd{A-*}-trained models underperform LA on Clotho. The differing retrieval behavior is therefore more consistent with differences in benchmark construction and caption style than with contamination.

\begin{table*}[t]
    \centering
    \caption{
    \textbf{Removing AF3-seen AudioCaps clips does not disproportionately affect \bvd{A-*}-trained models.}
    We report the mean change in audio-to-text (aR@5) and text-to-audio (tR@5) retrieval when evaluating on 607 AF3-clean AudioCaps clips relative to a fixed, randomly sampled control set of the same size. Negative values indicate lower performance on the AF3-clean subset. We additionally indicate whether each configuration was trained on \bvd{A-*} data and whether its retrieval training data include AudioCaps or AudioSet-derived data.
    }
    \label{tab:af3_decontamination}
    \begin{tabular}{lrrrr}
        \toprule
        \textbf{Training Data} &
        \textbf{aR@5} &
        \textbf{tR@5} &
        \textbf{BVD-A} &
        \textbf{AC/AS-derived} \\
        \midrule
        \bvd{A-1.7M}       & $-0.39$ & $-4.50$ & Yes & No  \\
        \bvd{A-10M}        & $-0.81$ & $-2.52$ & Yes & No  \\
        AC+CL+\bvd{A-1.7M} & $-2.40$ & $-3.44$ & Yes & Yes \\
        LA               & $+0.97$ & $-1.48$ & No  & No  \\
        AC+CL+LA         & $-2.98$ & $-3.52$ & No  & Yes \\
        AC+CL+LA+AS      & $-3.19$ & $-2.91$ & No  & Yes \\
        LA+AS            & $-3.49$ & $-3.85$ & No  & Yes \\
        \midrule
        BVD-A-trained (pooled)   & $-1.20$ & $-3.49$ & -- & -- \\
        Non-BVD-A (pooled)       & $-2.56$ & $-3.09$ & -- & -- \\
        \bottomrule
    \end{tabular}
\end{table*}

\begin{table*}[t]
    \centering
    \caption{
    \textbf{\bvd{A-1.7M} exhibits a broader caption-length distribution than the AudioCaps and Clotho test sets.}
    We report the percentage of captions falling into different caption-length intervals, measured in words.
    }
    \label{tab:caption_length_distribution}
    \begin{tabular}{lrrrrrrr}
        \toprule
        \textbf{Dataset} &
        \textbf{$<5$} &
        \textbf{5--10} &
        \textbf{10--15} &
        \textbf{15--20} &
        \textbf{20--25} &
        \textbf{25--30} &
        \textbf{$>30$} \\
        \midrule
        \bvd{A-1.7M}       & 2.0\%  & 34.7\% & 25.4\% & 16.0\% & 11.8\% & 6.0\% & 4.0\% \\
        AudioCaps (Test) & 10.6\% & 48.3\% & 24.3\% & 11.0\% & 4.4\%  & 1.3\% & 0.2\% \\
        Clotho (Test)    & 0.0\%  & 44.8\% & 44.6\% & 10.6\% & 0.0\%  & 0.0\% & 0.0\% \\
        \bottomrule
    \end{tabular}
\end{table*}

\subsection{Image-Text (Frame-based CLIP)}\label{sec:clip_appendix}

\subsubsection{Experimental setup}
For training CLIP models we used the OpenCLIP~\citep{ilharco_gabriel_2021_5143773} codebase. We trained models on different datasets with the same setup and hyperparameters outlined in \cref{appendix:training_hypers}.

\begin{table}[ht]
\centering
\caption{\textbf{Hyperparameters for CLIP ViT-B/32 and ViT-B/16.}}
\label{appendix:training_hypers}
\resizebox{\textwidth}{!}{
    \begin{tabular}{llcccccc}
    \toprule
    \textbf{Model} & \textbf{Samples Seen} & \textbf{LR Scheduler} & \textbf{Warmup Steps} & \textbf{LR} & \textbf{GPUs} & \textbf{Batch Size} \\
    \midrule
    \multirow{5}{*}{ViT-B/32} 
      & 12.8M  & cosine & 4000 & 0.001 & 16 GPUs A100 & 2048 \\
      & 30.7M  & cosine & 3000 & 0.001 & 16 GPUs A100 & 4096 \\
      & 64.0M  & cosine & 4000 & 0.002 & 16 GPUs A100 & 8192 \\
      & 128.0M & cosine & 4000 & 0.002 & 64 GPUs A100 & 8192 \\
      & 307.2M & cosine & 8000 & 0.002 & 64 GPUs A100 & 16384 \\
    \midrule
    \multirow{5}{*}{ViT-B/16} 
      & 12.8M  & cosine & 4000 & 0.001 & 16 GPUs A100 & 2048 \\
      & 30.7M  & cosine & 4000 & 0.002 & 16 GPUs A100 & 4096 \\
      & 64.0M  & cosine & 4000 & 0.002 & 16 GPUs A100 & 8192 \\
      & 128.0M & cosine & 6000 & 0.002 & 64 GPUs A100 & 8192 \\
      & 307.2M & cosine & 4000 & 0.002 & 64 GPUs A100 & 16384 \\
    \bottomrule
    \end{tabular}
}
\end{table}

For synthetic frame-caption generation, the \texttt{vLLM}~\citep{kwon2023efficient} framework was used, alongside a Ray~\citep{moritz2018ray} cluster to distribute the workload across nodes. We used 256 NVIDIA-A100 GPUs to efficiently produce captions for 300M frames. Since \bvd{I-300M} frames have a higher resolution than DataComp images ($720\,$px vs $256\,$px), we could not achieve the same throughput ($8.3$ img/s). The number of GPU hours for CLIP training and captioning is summarized in \cref{tab:gpu_hours}. The hyperparameters used for captioning can be found in \cref{tab:captioning_hyperparameters}.

\begin{table}[ht]
\centering
\begin{tabular}{ll}
\toprule
\textbf{Experiment} & \textbf{A100 Hours} \\
\midrule
CLIP Training & 2187 \\
Captioning & 12200 \\
\bottomrule
\end{tabular}
\caption{Total GPU hours consumed}
\label{tab:gpu_hours}
\end{table}

\begin{table}[h]
\centering
\begin{tabular}{ll}
\toprule
\textbf{Hyperparameter} & \textbf{Value} \\
\midrule
Temperature & 0.5 \\
Max Model Length & 4096 \\
Max Sequences & 16 \\
Max Tokens & 20 \\
\bottomrule
\end{tabular}
\caption{Frame-Level Captioning Hyperparameters}\label{tab:captioning_hyperparameters}
\end{table}

\subsubsection{Captioning}\label{sec:clip_captioning}
We select a VLM that balances quality and efficiency, and then apply it to generate frame-level captions. We validate the effectiveness of our frame captions through downstream performance on standard benchmarks and explore the impact of caption length on the performance.

\paragraph{Selecting a captioning model}
We considered models for image captioning that balance quality and throughput, selecting the seven top-performing models from the OpenVLM leaderboard~\citep{duan2024vlmevalkit} as our starting point.
Due to practical constraints on overall compute and GPU memory, we limited our selection to VLMs with fewer than 4B parameters.
The resulting candidate pool is provided in \cref{tab:captioners}.
We use CLIPScore~\citep{hessel2021clipscore} as a proxy to measure the quality of generated captions.
Specifically, we calculate CLIPScore using OpenAI~CLIP~ViT-L/14@336px~\citep{openai2021clip336} on a representative subset of generated captions for 100k images from DataComp-1B~\citep{gadre2023datacomp}.
We select DeepSeek-VL2-tiny~\citep{wu2024deepseekvl2mixtureofexpertsvisionlanguagemodels} for captioning our dataset, as it yielded the highest CLIPScore and second-best throughput. %
\begin{table}[t]
    \centering
        \caption{\textbf{Candidate captioner models}. We choose DeepSeek-VL2-tiny, which achieves the \textbf{highest} CLIPScore and the \uline{second-highest} throughput.}
        \label{tab:captioners}
    \resizebox{\textwidth}{!}{%
        \sisetup{
          detect-weight=true,
        }
        \begin{tabular}{lllrS[table-format=1.2]S[table-format=2.2]}
        \toprule
        & & & & & \textbf{Throughput} \\
        \textbf{Model} & \textbf{Language Model} & \textbf{Vision Model} & \textbf{Params} & \textbf{CLIPScore}  & {in img/s} \\
        \midrule
        InternVL2.5-1B~\citep{chen2024internvlscalingvisionfoundation} & Qwen-2.5-0.5B & InternViT-300M-v2.5 & 1\,B & 0.43 & \bfseries 10.41 \\
        InternVL2.5-2B-MPO~\citep{chen2024internvlscalingvisionfoundation} & InternLM2.5-1.8B & InternViT-300M-v2.5 & 2\,B & 0.27 & 6.41 \\
        InternVL2.5-2B~\citep{chen2024internvlscalingvisionfoundation} & InternLM2.5-1.8B & InternViT-300M-v2.5 & 2\,B & 0.21 & 6.21 \\
        SmolVLM-Instruct~\citep{marafioti2025smolvlm} & SmolLM2-1.7B & SigLIP-400M & 2.3\,B & 0.50 & 2.71 \\
        \rowcolor{black!10}DeepSeek-VL2-tiny~\citep{wu2024deepseekvl2mixtureofexpertsvisionlanguagemodels} & DeepSeekMoE-3B & SigLIP-400M & 3.4\,B & \bfseries 0.62 & \Uline{8.20} \\
        Qwen2.5-VL-3B-Instruct~\citep{bai2025qwen25vl} & Qwen2.5-3B & QwenViT & 3.75\,B & 0.55 & 2.46 \\
        Phi-3.5-vision-instruct~\citep{abdin2024phi} & Phi-3.5-mini-instruct & OpenAI CLIP ViT-L/14@336px & 4.15\,B & \Uline{0.59} & 3.74 \\
        \bottomrule
        \end{tabular}
   }
\end{table}

\paragraph{Validation}
To assess caption quality, we recaption subsets of DataComp-1B~\citep{gadre2023datacomp} using DeepSeek-VL2-tiny and train CLIP ViT-B/32.
For validation, we consider the zero-shot image classification performance on ImageNet-1k~\citep{deng2009imagenet}, ImageNet-R~\citep{hendrycks2021many}, and ImageNet-Sketch~\citep{wang2019learning} alongside image retrieval performance on COCO Captions~\citep{lin2014microsoft}.
The results are summarized in \cref{tab:caption-validation}.
While our automatically generated captions lead to a drop in classification accuracy (\SI{36}{\%} for ImageNet, \SI{11}{\%} and \SI{16}{\%} for ImageNet-R and ImageNet-Sketch respectively at the largest data scale), image retrieval performance increases by \SI{27}{\%} compared to the original captions.
Overall, we find the caption quality acceptable considering the reduced annotation cost and increased dataset scale.

\begin{table}[h]
    \centering
    \caption{
        \textbf{Validation of caption quality}.
        We report the performance of CLIP ViT-B/32 trained on subsets of DataComp-1B~\citep{gadre2023datacomp} with their original captions, our automatically generated captions (Recap), and length-constrained captions with only 7 words on average (Short).
    }
    \label{tab:caption-validation}
    \resizebox{\textwidth}{!}{%
        \begin{tabular}{rS[table-format=1.3]S[table-format=1.3]S[table-format=1.3]S[table-format=1.3]S[table-format=1.3]S[table-format=1.3]S[table-format=1.3]S[table-format=1.2]S[table-format=1.2]S[table-format=1.2]}
        \toprule
        & \multicolumn{7}{c}{\textbf{Zero-Shot Classification} Acc@1} & \multicolumn{3}{c}{\textbf{Img-Retrieval} Recall@5} \\
        \cmidrule(lr){2-8}
        & \multicolumn{3}{c}{\textbf{ImageNet-1k}} & \multicolumn{2}{c}{\textbf{ImageNet-R}} & \multicolumn{2}{c}{\textbf{ImageNet-Sketch}} & \multicolumn{3}{c}{\textbf{COCO Captions}} \\
        \cmidrule(lr){2-4} \cmidrule(lr){5-6} \cmidrule(lr){7-8} \cmidrule(lr){9-11}
        \textbf{Samples} & \text{Original} & \text{Recap} & \text{Short} & \text{Original} & \text{Recap} & \text{Original} & \text{Recap} & \text{Original} & \text{Recap} & \text{Short} \\
        \midrule
        12.8\,M & 0.10 & 0.09 & 0.10  & 0.12 & 0.15 & 0.04 & 0.06 & 0.10  & 0.21  & 0.19 \\
        30.7\,M & 0.20  & 0.15  & 0.17 & 0.21 & 0.24 & 0.10 & 0.13 & 0.18  & 0.31  & 0.28 \\
         128\,M & 0.40  & 0.23  & \textendash & 0.41  & 0.36 & 0.25  & 0.21  & 0.33  & 0.42  & \textendash \\
        \bottomrule
        \end{tabular}
    }
\end{table}

\paragraph{Caption length}
We explore different input prompts for captioning and consequently the impact of different resulting caption lengths on downstream model performance. Specifically, we use the following prompt as our default choice: 
\hl{\texttt{
Provide a very coarse brief single line of caption for the image.}}
We compare this to using the following prompt, which resulted in caption lengths limited to around 7 words on average (short): \hl{\texttt{Provide a very coarse brief single line of caption for the main object in the image. Don't worry about the details, just a very high-level description. The description should be as short as possible - 1-3 words.}}
The impact on model performance of these short captions is also included in \cref{tab:caption-validation}.
While we observe a small improvement in classification accuracy on 30.7M training samples, we ultimately decide against artificially constraining the caption length.

\subsubsection{Frame filtering}
\cref{tab:extracted_frames} shows the number of frames extracted and the corresponding percentage of videos.

\begin{table}[h]
\centering
\caption{Number of extracted frames per video using our filtering pipeline and corresponding percentage of videos.}
\label{tab:extracted_frames}
\begin{tabular}{lc}
\toprule
\textbf{Extracted frames} & \textbf{Percentage of videos} \\
\midrule
1   & 12.93\% \\
2   & 7.96\%  \\
3   & 6.31\%  \\
4   & 5.36\%  \\
5   & 4.74\%  \\
6   & 4.21\%  \\
7   & 3.79\%  \\
8   & 3.44\%  \\
9+  & 51.27\% \\
\bottomrule
\end{tabular}
\end{table}

\section{Why Strong Retrieval Does Not Translate to ImageNet-1k Accuracy}
\label{app:caption_analysis}

Here we want to expand our analysis of the surprisingly strong retrieval performance (and suboptimal ImageNet-1k classification accuracy) on the filtered and captioned \bvd{I-300M} frames as shown in \cref{tab:exp_results_clip}. \cref{tab:synthetic_captions_results} shows that models trained on subsets of \bvd{I-300M} also achieve strong Flickr30k image-to-text retrieval performance (up to 0.87 R@5). \\
Additionally, we analyze both the caption length distribution and the coverage of ImageNet-1k class names in two 300M-sample subsets: (i) synthetic captions in \bvd{I-300M} and (ii) original (alt-text) captions from DataComp.

\begin{figure}[t]
    \centering
    \includegraphics[width=0.45\linewidth]{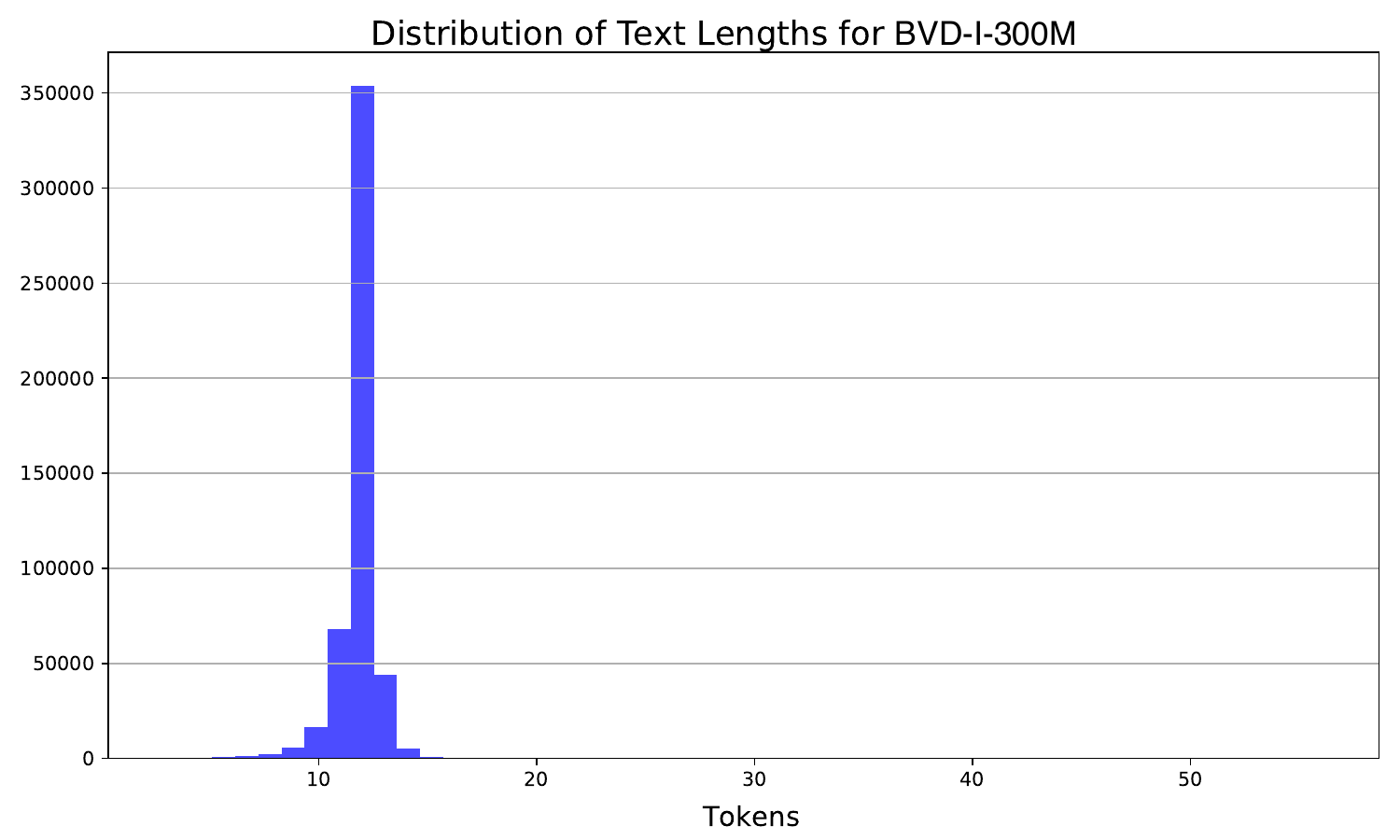}
    \includegraphics[width=0.45\linewidth]{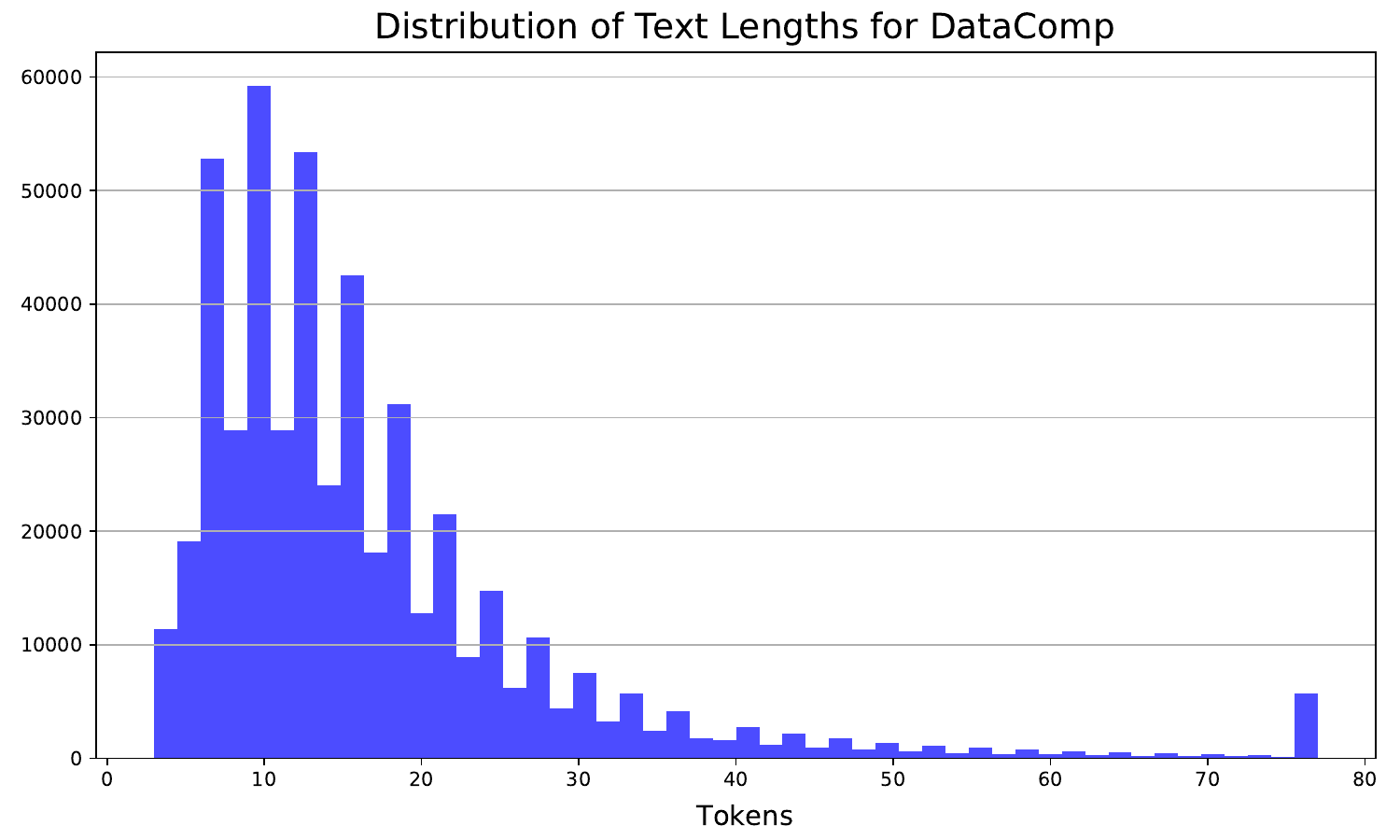}
    \caption{Caption length distribution for \bvd{I-300M} synthetically captioned subset (left) and the 300M DataComp subset (right). \bvd{I-300M} captions are longer with a narrower distribution, while DataComp captions show a broader distribution dominated by short captions.}
    \label{fig:caption_length}
\end{figure}

\paragraph{Caption length distribution}
\cref{fig:caption_length} shows the frequency distribution of the number of tokens per caption. Captions in \bvd{I-300M} exhibit a noticeably \emph{narrower} distribution, with longer captions dominating. In contrast, captions in the DataComp subset display a \emph{broader} distribution, with a substantial fraction of short captions. This indicates that synthetic \bvd{I-300M} captions resemble verbose, COCO-style descriptions, while real DataComp captions reflect the more diverse and often concise nature of web text.

\begin{figure}[t]
    \centering
    \includegraphics[width=0.9\linewidth]{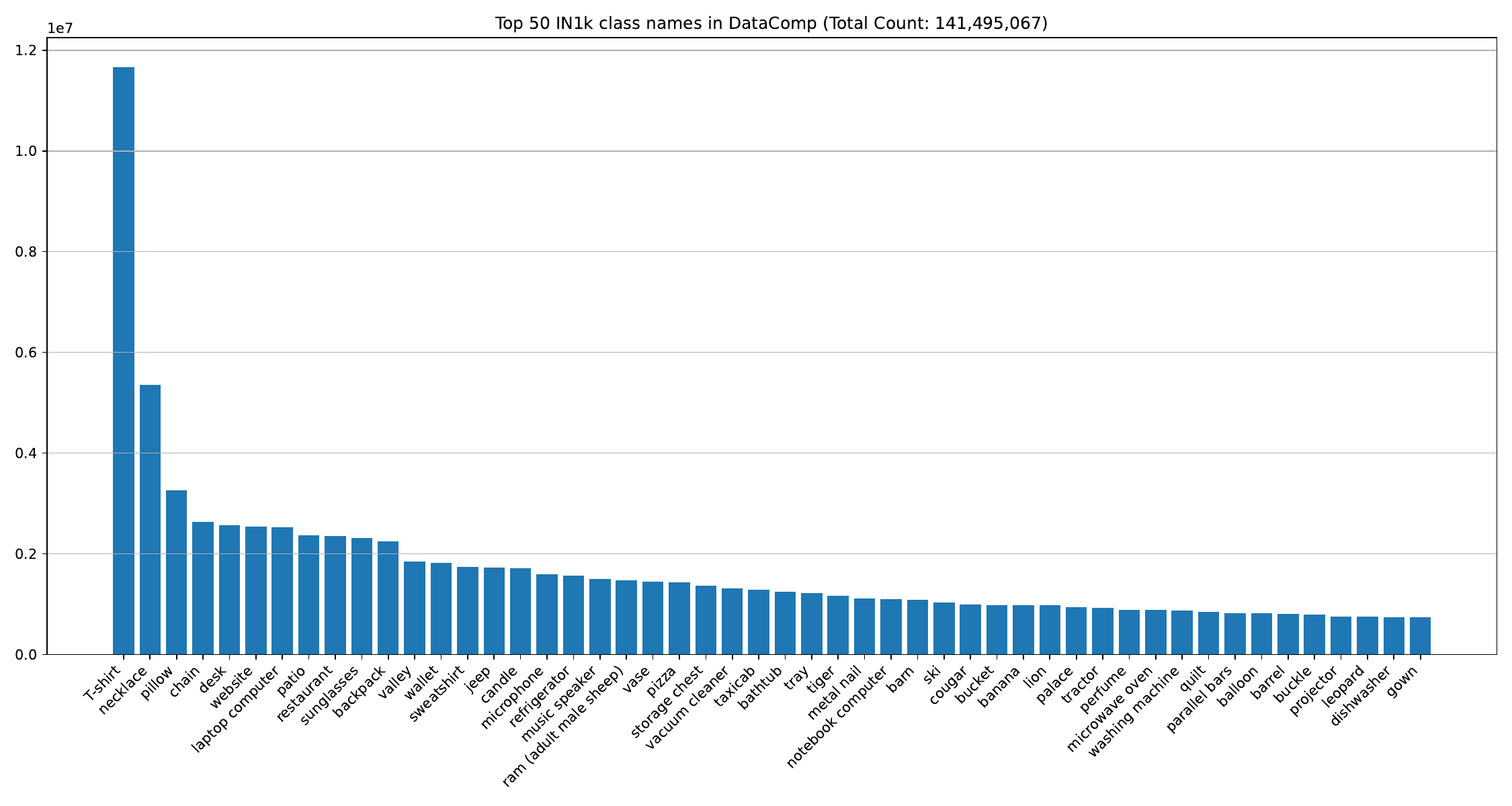}
    \caption{Top-50 most frequent ImageNet-1k class names appearing in the 300M DataComp captions. Real captions from DataComp strongly align with ImageNet-1k semantic categories, exhibiting 141M total class-name occurrences.}
    \label{fig:datacomp_classes}
\end{figure}

\begin{figure}[t]
    \centering
    \includegraphics[width=0.9\linewidth]{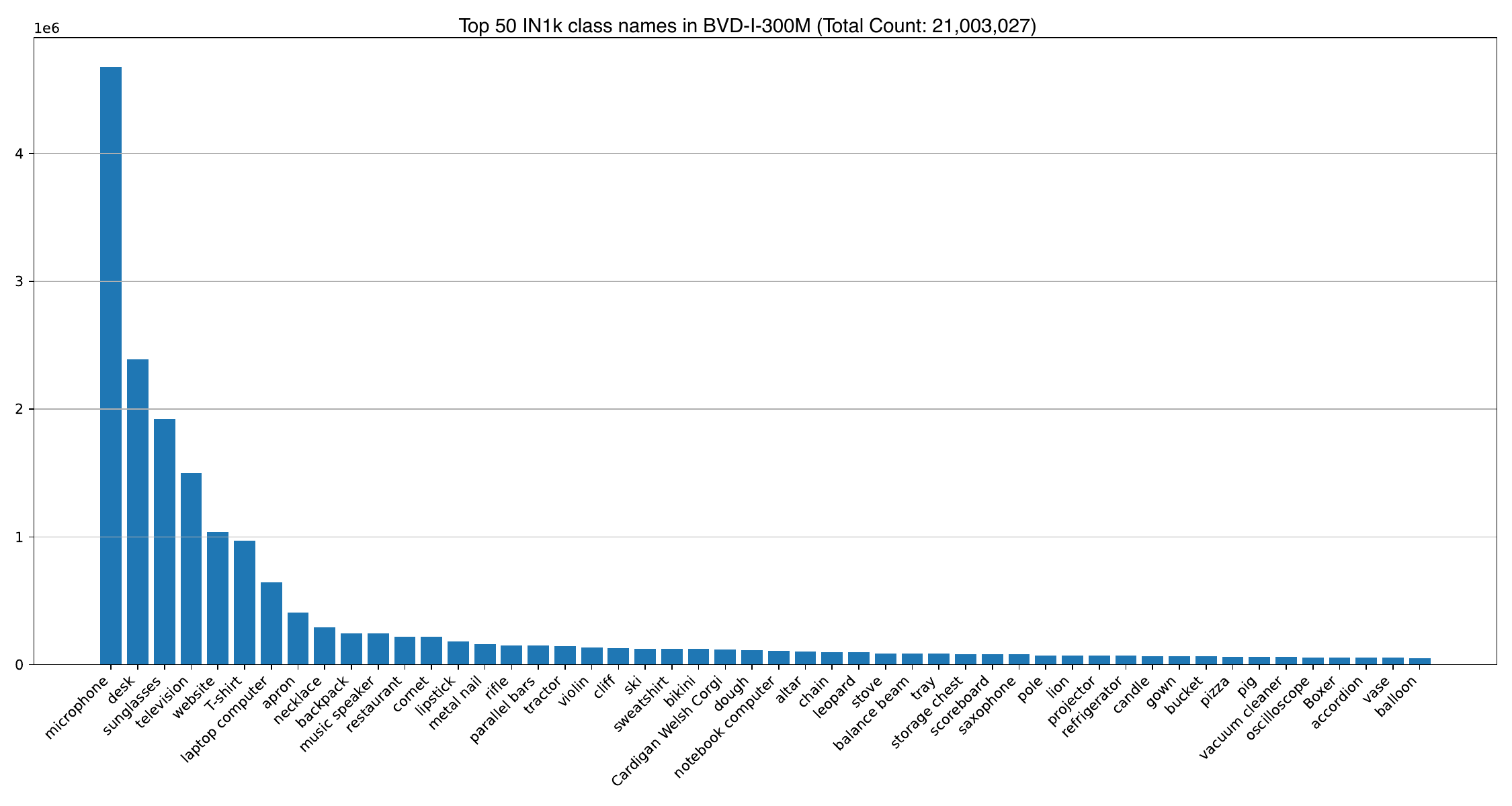}
    \caption{Top-50 most frequent ImageNet-1k class names appearing in the \bvd{I-300M} synthetic captions. \bvd{I-300M} captions contain 21M total class-name occurrences, substantially fewer than DataComp, consistent with the observed differences in ImageNet-1k zero-shot performance.}
    \label{fig:ovid_classes}
\end{figure}

\paragraph{ImageNet-1k class-name coverage}
We further count the occurrences of ImageNet-1k class names (based on the corresponding synsets) in the two subsets. We identify approximately 21M class-name mentions in the \bvd{I-300M} captions, compared to 141M in the DataComp captions. \cref{fig:datacomp_classes} and \cref{fig:ovid_classes} show the top-50 most frequent ImageNet-1k class names for the respective datasets.

This substantial difference in class-name coverage provides a plausible explanation for the weaker ImageNet-1k zero-shot classification performance of models trained on \bvd{I-300M} subsets (see \cref{tab:synthetic_captions_results}). Real captions from DataComp appear to be more strongly aligned with the semantic structure of ImageNet-1k, containing a far greater density and diversity of class-associated terminology.

Taken together, these findings suggest that caption length or fluency alone is insufficient for improving downstream transfer. Instead, explicit alignment with class-centric semantics, by, for instance, increasing the inclusion of ImageNet-1k class names or concept-oriented vocabulary during synthetic caption generation may be a more effective strategy for boosting zero-shot classification performance on ImageNet-like benchmarks~\citep{ghosh2026concept}.

\begin{table}[h!]
\centering
\caption{Image-text retrieval performance (R@5) on Flickr30k and MS-COCO for models trained on different 300M-30M subsets of \bvd{I-300M}, DataComp, and Re-LAION. Scaling laws for \bvd{I-300M} persist on different models.}
\label{tab:synthetic_captions_results}
\resizebox{\textwidth}{!}{
    \begin{tabular}{llrcc}
    \toprule
    \textbf{Model} & \textbf{Dataset} & \textbf{Samples} & \textbf{Flickr30k I$\rightarrow$T R@5} & \textbf{MS-COCO I$\rightarrow$T R@5} \\
    \midrule
    ViT-B/16-text-plus & \bvd{I-300M} & 300M & 0.87 & 0.63 \\
    ViT-B/16-text-plus & \bvd{I-300M} & 128M & 0.82 & 0.56 \\
    ViT-B/32           & \bvd{I-300M} & 300M & 0.82 & 0.56 \\
    ViT-B/16-text-plus & Re-LAION  & 300M & 0.80 & 0.53 \\
    ViT-B/16-text-plus & \bvd{I-300M} & 64M  & 0.79 & 0.50 \\
    ViT-B/16-text-plus & DataComp & 300M & 0.75 & 0.52 \\
    ViT-B/32           & \bvd{I-300M} & 128M & 0.75 & 0.48 \\
    ViT-B/32           & Re-LAION  & 300M & 0.72 & 0.46 \\
    ViT-B/16-text-plus & Re-LAION  & 128M & 0.71 & 0.45 \\
    ViT-B/16-text-plus & \bvd{I-300M} & 30M  & 0.67 & 0.40 \\
    ViT-B/32           & \bvd{I-300M} & 64M  & 0.67 & 0.40 \\
    ViT-B/32           & DataComp & 300M & 0.66 & 0.45 \\
    ViT-B/16-text-plus & DataComp & 128M & 0.65 & 0.43 \\
    ViT-B/32           & Re-LAION  & 128M & 0.61 & 0.38 \\
    ViT-B/32           & DataComp & 128M & 0.55 & 0.37 \\
    ViT-B/16-text-plus & DataComp & 64M  & 0.55 & 0.36 \\
    ViT-B/32           & \bvd{I-300M} & 30M  & 0.54 & 0.31 \\
    ViT-B/32           & Re-LAION  & 64M  & 0.51 & 0.30 \\
    ViT-B/32           & DataComp & 64M  & 0.44 & 0.29 \\
    ViT-B/16-text-plus & DataComp & 30M  & 0.38 & 0.26 \\
    ViT-B/32           & Re-LAION  & 30M  & 0.37 & 0.22 \\
    ViT-B/32           & DataComp & 30M  & 0.32 & 0.21 \\
    \bottomrule
    \end{tabular}
}
\end{table}

\section*{Disclaimer for use of LLMs}
Large Language Models (LLMs) were used for language polishing and proofreading to improve clarity and readability, as well as for caption generation (see \cref{sec:captioning}). In addition, LLM-based coding agents were used to support coding tasks. All ideas, experimental design, and scientific content originated from the authors.

\end{document}